%% file: main.tex
\documentclass[ettersize,journal]{IEEEtran}
\usepackage[utf8]{inputenc} % allow utf-8 input
\usepackage[T1]{fontenc}    % use 8-bit T1 fonts
\usepackage{hyperref}       % hyperlinks
\usepackage{url}            % simple URL typesetting
\usepackage{booktabs}
\usepackage{textcomp}
\usepackage{balance}
\usepackage{amsfonts}       % blackboard math symbols
\usepackage{nicefrac}       % compact symbols for 1/2, etc.
\usepackage{microtype}      % microtypography
\usepackage[ruled,linesnumbered]{algorithm2e}      % algorithm
\usepackage{subcaption}
\usepackage{mathtools}
\usepackage{bm}
\usepackage{xcolor}         % colors
\usepackage[
  separate-uncertainty = true,
  multi-part-units = repeat
]{siunitx}

\usepackage[font={small}]{caption} %caption font smaller

\newcommand{\R}{{\mathbb{R}}}
\newcommand{\E}{{\mathbb{E}}}
\DeclareMathOperator*{\argmax}{arg\,max}

\usepackage[square,numbers]{natbib}
\renewcommand{\bibsection}{\subsubsection*{\bibname}}

\usepackage{eqparbox}
\newcommand{\algorithmiccomment}[1]{\hfill\eqparbox{COMMENT}{\# #1}}

\newcommand{\s}{{\bm{s}}}
\newcommand{\x}{{\bm{x}}}

\newcommand{\y}{{\bm{y}}}

\newcommand{\calL}{{\mathcal{L}}}
\usepackage{amsthm}
\newtheorem{prop}{Proposition}
\renewcommand{\appendixname}{Supplementary Material}

\usepackage[misc,geometry]{ifsym}
\newcommand{\corrAuthor}{$^{\textrm{\Letter}}$}
\newcommand{\coFirstAuthor}{$^{\textrm{\#}}$}

\ifCLASSINFOpdf
\else
\fi
\begin{document}
\title{Supervising the Decoder of Variational\\ Autoencoders to Improve Scientific Utility}

\author{Liyun Tu\coFirstAuthor$^,$ \corrAuthor,
        Austin Talbot\coFirstAuthor,
        Neil M. Gallagher,
        and David E. Carlson% <-this % stops a space
\thanks{\coFirstAuthor equal contribution, \corrAuthor corresponding author.}
\thanks{L. Tu is with the School of Artificial Intelligence, Beijing University of Posts and Telecommunications, Beijing, 100876, China. L. Tu is the corresponding author. Email: tuliyun@gmail.com. This work was done when L. Tu was working as a postdoctoral associate at the Department of Civil and Environmental Engineering in Duke University, Durham, NC 27708, USA. } % <-this % stops a space
\thanks{A. Talbot is with the Department of Psychiatry and Behavioral Sciences, Stanford University, Palo Alto, CA 94305, USA. Email: abt23@stanford.edu.} % <-this % stops a space
\thanks{N. M. Gallagher is with the Department of Psychiatry, Weill Cornell Medical College, New York, NY 10065, USA. Email: nga4004@med.cornell.edu.} % <-this % stops a space
\thanks{D. E. Carlson is with the Department of Biostatistics and Bioinformatics and the Department of Civil and Environmental Engineering, Duke University, Durham, NC 27708, USA. Email: david.carlson@duke.edu.} % <-this % stops a space
% \thanks{Manuscript received April 19, 2005; revised August 26, 2015.}
%\thanks{Manuscript received **, 2021; revised **, 2021.}
}


\markboth{}%
{Shell \MakeLowercase{\textit{et al.}}: Supervising the Decoder of Variational\\ Autoencoders Improves Scientific Utility}



\maketitle

\begin{abstract}
\input{./texfiles/abstract}
\end{abstract}


% \fi
\IEEEpeerreviewmaketitle

% \section{Introduction}
% % The very first letter is a 2 line initial drop letter followed
% % by the rest of the first word in caps.
% % 
% % form to use if the first word consists of a single letter:
% % \IEEEPARstart{A}{demo} file is ....
% % 
% % form to use if you need the single drop letter followed by
% % normal text (unknown if ever used by the IEEE):
% % \IEEEPARstart{A}{}demo file is ....
% % 
% % Some journals put the first two words in caps:
% % \IEEEPARstart{T}{his demo} file is ....
% % 
% % Here we have the typical use of a "T" for an initial drop letter
% % and "HIS" in caps to complete the first word.
% for IEEE journal papers produced under \LaTeX\ using
% IEEEtran.cls version 1.8b and later.
% % You must have at least 2 lines in the paragraph with the drop letter
% % (should never be an issue)
% I wish you the best of success.
\input{texfiles/introduction}

 

\input{texfiles/relatedWork}

\input{texfiles/methods}

\input{texfiles/extensions}

\input{texfiles/results}

\input{texfiles/conclusions}

\section*{Acknowledgments}

Research reported in this manuscript was supported by the National Institute of Biomedical Imaging and Bioengineering and the National Institute of Mental Health through the National Institutes of Health BRAIN Initiative under Award Number R01EB026937.

The contents of this manuscript are solely the responsibility of the authors and do not necessarily represent the official views of any of the funding agencies or sponsors.
\\

%ues IEEE style title/heading
\section*{References}
%remove the default title/heading: "References".
\renewcommand{\bibsection}{}
\renewcommand{\bibfont}{\small}
\bibliography{main}
\renewcommand{\bibfont}{\normalsize}


\begin{IEEEbiography}
    [{\includegraphics[width=1in,height=1in,clip,keepaspectratio]{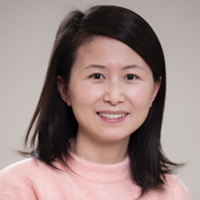}}]{Liyun Tu}received her PhD degree in Computer Science and Technology as well as an MS and BSE in software engineering, from Chongqing University, China. She worked as a postdoctoral associate in Duke University (2019-2021) and in Children’s National Hospital (2017-2019), USA. Her research interests include medical image analysis, machine learning, and statistical modeling.
\end{IEEEbiography}

\begin{IEEEbiography}
    [{\includegraphics[width=1in,height=1in,clip,keepaspectratio]{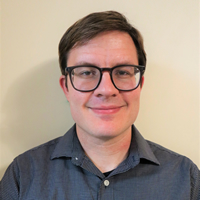}}]{Austin Talbot}received a BS degree in applied and computational mathematics from Brigham Young University in 2015. He received his MS and PhD in statistical science from Duke University in 2019 and 2020 respectively. He is currently a postdoctoral scholar at Stanford University in the department of Psychiatry and Behavioral Science. His research interests include robust statistics, supervised learning, latent variable models, and psychiatric stimulation techniques.
\end{IEEEbiography}

\begin{IEEEbiography}
    [{\includegraphics[width=1in,height=1in,clip,keepaspectratio]{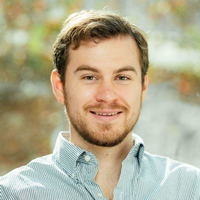}}]{Neil M. Gallagher}
received the BSE degree in biomedical engineering from Duke University in 2013, the MS degree in electrical engineering from Stanford University in 2015, and the PhD degree in neurobiology from Duke University in 2021. He is currently a postdoctoral researcher at the Weill Cornell Medical Center in New York City. His research interests include neural time series analysis, mental state decoding, and closed-loop neural stimulation.
\end{IEEEbiography}

\begin{IEEEbiography}
    [{\includegraphics[width=1in,height=1in,clip,keepaspectratio]{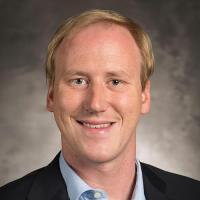}}]{David E. Carlson}
received his BSE, MS, and PhD from Duke University in Durham, NC in 2010, 2014, and 2015, respectively.  He previously completed postdoctoral training at Columbia University in New York, NY.  He is currently an Assistant Professor of Civil and Environmental Engineering, Biostatistics and Bioinformatics, Electrical and Computer Engineering, and Computer Science.  His research interests include machine learning methods and applications.
\end{IEEEbiography}

% or

%\onecolumn
\clearpage
%\onecolumn
\appendix

%\begin{multicols}{2}
\input{texfiles/appendix}

%\end{multicols}


\end{document}

%% file: texfiles/abstract.tex
Probabilistic generative models are attractive for scientific modeling because their inferred parameters can be used to generate hypotheses and design experiments. This requires that the learned model provides an accurate representation of the input data and yields a latent space that effectively predicts outcomes relevant to the scientific question. Supervised Variational Autoencoders (SVAEs) have previously been used for this purpose, as a carefully designed decoder can be used as an interpretable generative model of the data, while the supervised objective ensures a predictive latent representation. Unfortunately, the supervised objective forces the encoder to learn a biased approximation to the generative posterior distribution, which renders the generative parameters unreliable when used in scientific models. This issue has remained undetected as reconstruction losses commonly used to evaluate model performance do not detect bias in the encoder. We address this previously-unreported issue by developing a second-order supervision framework (SOS-VAE) that updates the decoder parameters, rather than the encoder, to induce a predictive latent representation. This ensures that the encoder maintains a reliable posterior approximation and the decoder parameters can be effectively interpreted. We extend this technique to allow the user to trade-off the bias in the generative parameters for improved predictive performance, acting as an intermediate option between SVAEs and our new SOS-VAE. We also use this methodology to address missing data issues that often arise when combining recordings from multiple scientific experiments. We demonstrate the effectiveness of these developments using synthetic data and electrophysiological recordings with an emphasis on how our learned representations can be used to design scientific experiments.

%that learns a latent representation predictive of the outcome while maintaining a reliable generative interpretation. We extend this technique to allow the user to trade-off some bias in the generative parameters for improved predictive performance, acting as an intermediate option between SVAEs and our new SOS-VAE. We also use this methodology to address missing data issues that often arise when combining recordings from multiple scientific experiments. We demonstrate the effectiveness of these developments using synthetic data and electrophysiological recordings with an emphasis on how our learned representations can be used to design scientific experiments.

%% file: texfiles/introduction.tex
\section{Introduction}
\label{sec:intro}

\IEEEPARstart{D}{eveloping} interpretable and explainable generative models has long been an integral area of machine learning and Bayesian modeling \citep{Blei2003LatentAllocation,Perfors2011ADevelopment,Bonawitz2010DeconfoundingModels,Gallagher2017Cross-spectralAnalysis}. 
Generative models have great scientific utility in developing testable hypotheses \citep{Carlson2017Dynamically-timedPathway,Hultman2018Brain-wideVulnerability}. Specfically, an interpretable relationship between a generative model representation and the observed covariates can provide insight as to how to develop causal scientific experiments \cite{Mague2020Brain-wideState,block2020prenatal}. 

% I think this would be better in the discussion - it isn't the main question we're asking so it's confusing to have it in the intro - Neil
%This insight is obtained from the generative parameters and relies on their clear connection to the latent variables via the posterior defined by the generative model. While ideally these latent variables would be obtained from the true posterior of the generative model, tractable inference often requires that this posterior be approximated, such as in Variational Autoencoders \citep{Kingma2013} (VAEs). However, the clear relationship between the true posterior and the variational approximation has allowed for these models to maintain their scientific utility \citep{Srivastava2017AutoencodingModels,Speiser2017FastAutoencoders,Burkhardt2019DecouplingModel,Talbot2020SupervisedActivity}. 

An interpretable relationship between the observed covariates and model representation, while necessary, is often not sufficient in scientific settings. Many scientific applications require that the learned representation also be predictive of an auxiliary variable \citep{Jolliffe1982}. In the neuroscience research motivating this work, this is often a behavioral outcome \citep{mague2020brain}, a genetic phenotype \citep{jiang_generative_2020}, or the presence of some disorder or disability \citep{zhao_constructing_2017}. Obtaining a predictive latent representation allows for researchers to identify patterns relevant to the auxiliary variables and establish causality through experimental modification \citep{Vu2018ANeuroscience}. Unfortunately, exclusively generative models often fail to represent the desired auxiliary variable, and are typically dominated by other irrelevant sources of variation. Returning to our motivating work as an example, neural dynamics associated with motion \citep{Khorasani2019} and even blinking \citep{Joyce2004} are often substantially  stronger than the auxiliary variables of interest.

One class of generative models called Variational Autoencoders (VAEs) can be encouraged to yield a predictive latent representation by including a supervision loss during training. This yields a standard Supervised Variational Autoencoder (SVAE), which has been commonly used in the machine learning community \citep{Rasmus2015Semi-SupervisedNetworks,Raiko2018HowNetworks,Pu2016}. In those applications, the reconstruction loss of the generative model has been motivated as an effective method of improving predictive models, as the reconstructive loss is theoretically and empirically an effective regularization technique \citep{Le2018}. Based on this work it might seem that SVAEs satisfy both the generative and predictive criteria for scientific utility.

%Because of this, an SVAE seems to enjoy the benefits of both worlds, a generative interpretation that improves predictive generalization.

Unfortunately, the supervision loss in a standard SVAE (referred to as SVAE in the remaining sections) biases the encoder away from approximating the true posterior distribution of latent representations. We demonstrate this bias by analyzing the fixed points of the SVAE objective. The bias caused by the inclusion of the supervision loss has not been noted in previous work, in part because those applications primarily focused on obtaining a predictive model. When the generative aspects of SVAEs are viewed merely as a convenient and effective regularization technique, the scientific utility of the decoder parameters is irrelevant. However, this bias in the variational objective can have a profoundly negative impact on causal scientific experiments designed using the decoder parameters. In these applications, the model will give misleading conclusions on how modification of the observed covariates will influence the auxiliary variable via the latent representation. As our motivating work uses these generative decoder models as a means for informing scientific manipulations, addressing this issue is critical. 
\begin{figure}[t]
    \centering
    \includegraphics[width=0.99\columnwidth]{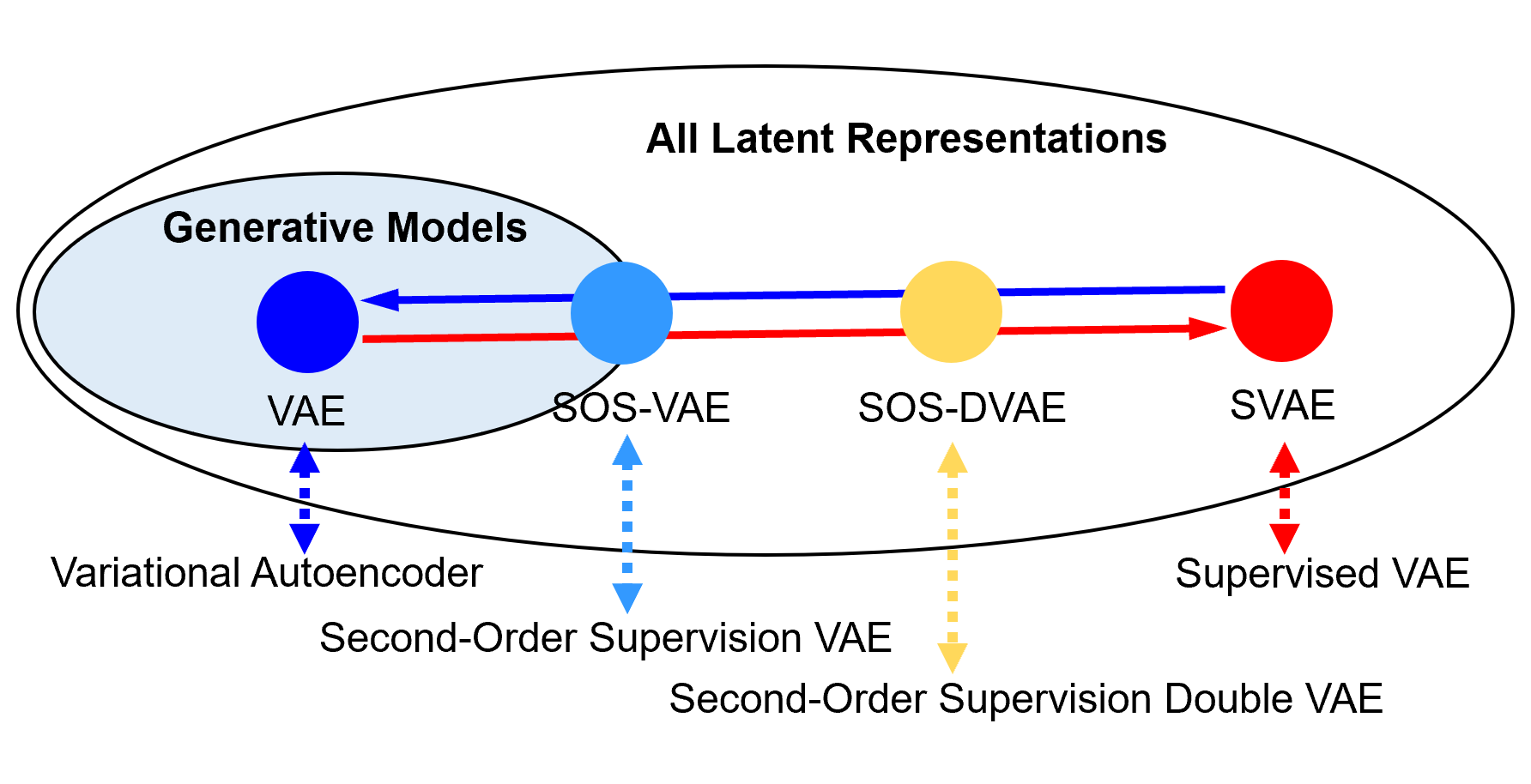}
    \caption{Relationship among the proposed models (SOS-VAE and SOS-DVAE, detailed in Sections \ref{sec:Methods} and \ref{sec:Extensions}, respectively), a generative model (VAE) and a predictive model (SVAE). Blue indicates a model pursues a lower reconstruction error (better generative performance), and red denotes that a model pursues a higher predictive performance. Both VAE and SOS-VAE try to find latent representations that only use information in the proper generative model family, with SOS-VAE trying to find a representation that is good at prediction within the model family. We note that the learned latent representation from SVAEs can be far from the generative model family, as we empirically show in Section \ref{sec:results}, and SOS-DVAE allows us to vary between optimal predictive performance and faithfulness to the generative model.
    \label{fig:sketch}}
\end{figure}

We remove this previously unobserved bias in SVAEs by developing a novel optimization framework using second-order gradient techniques \citep{Finn2017Model-agnosticNetworks,Liu2019Self-SupervisedLearning}. This Second-Order Supervision (SOS) framework maintains the interpretation of the variational encoder as an unbiased approximation to the posterior while inducing the latent representation to be predictive of an auxiliary variable (Fig. \ref{fig:sketch}). This approach yields a model that possesses both properties required for scientific utility, a predictive latent space and an unbiased posterior approximation. We empirically demonstrate these properties on a real dataset which uses Local Field Potentials (LFPs) recordings \citep{Gallagher2017Cross-spectralAnalysis, Carlson2017Dynamically-timedPathway} to predict a behavioral trait. We show that our framework yields more accurate predictions when compared to an exclusively generative model \citep{Plummer2014} while maintaining a more accurate variational approximation of the posterior when compared to an SVAE.

%We then develop two scientifically useful extensions of the initial framework. 
Previous work has shown that interpretable generative models can handicap a predictive objective as compared to a flexible neural network \citep{Hahn2013} due to the restrictive constraints imposed by the model family. %, as the predictions must follow the chosen model family, which is typically simpler than a complex neural network.
%, particularly relative to the flexible neural networks commonly used in variational approximations.
Because of this, our SOS framework can cause degraded predictive performance relative to an SVAE when the variational approximation is unconstrained by the generative model. % that is unconstrained by the generative model.
While sometimes this degradation is scientifically necessary and unavoidable, often a faint amount of bias is acceptable to obtain substantial gains in predictive ability. Our first extension provides a means to relax the constraint on the variational encoder, allowing it to maintain high predictive ability with minimal bias in the approximation. Using Kullback-Liebler (KL) divergences \citep{Kingma2013}, we demonstrate on both synthetic and real data examples that this regularization approach improves predictive performance for a given level of posterior discrepancy compared to an SVAE-based model. We denote this method as Second-Order Supervision Double Variational Autoencoder (SOS-DVAE), and conceptually show the relationship to the other models in Fig. \ref{fig:sketch}. We additionally show that this framework can be helpful in the context of missing data.

%Second, we develop a framework with multiple encoders to stitch multiple datasets together through the generative model.  This mimics the idea of ``shotgun sampling'' of scientific data where each dataset has a different subset of observed values or locations \citep{soudry2015efficient}. We show that our novel approach yields dramatically improved predictive performance relative to SVAEs when used in this fashion.

Thus, our work provides three major contributions to the field. First, we identify bias in SVAEs using a novel fixed point analysis, and show this bias has a detrimental impact on generative reconstruction. Second, we develop inference techniques to reduce or remove this bias in a computationally-feasible manner resulting in a more realistic generative model. Finally, we demonstrate how the proposed models (SOS-VAE, SOS-DVAE) can be applied in scientific settings by isolating brain networks of predictive of traits of interest. 

%improves predictive performance drastically over an SVAE method.

%% file: texfiles/relatedWork.tex
\section{Related Work}
\label{sec:related}

\textbf{Joint Factor Modeling.} Joint modeling, such as probabilistic supervised principal component analysis \citep{Yu2006} or supervised Gaussian processes \citep{Gao2011SupervisedReduction}, assumes that the observations and the outcome are independent given a latent representation. Once the prior distribution of the latent factors and the conditional distributions have been defined, statistical estimation is straightforward by maximum likelihood or Bayesian methods \citep{Bhattacharya2011}. However, it has been demonstrated that joint models suffer under model misspecification, particularly when the number of learned factors is less than the true latent dimensionality \citep{Hahn2013,Talbot2020SupervisedActivity}. Furthermore, the variance of the outcomes is often small relative to the variance of the observations, which leads to the outcome being poorly characterized \citep{Jolliffe1982}.

%Work on SAEs and generalization.
\textbf{Supervised Autoencoders.} To address some of these limitations, SAEs can be used to effectively approximate standard factor models while including a predictive term.  SAEs have been shown to ameliorate some concerns about model misspecficiation \citep{Talbot2020SupervisedActivity}. SAEs have been shown to give gains in many predictive applications \citep{Pu2016,Rasmus2015Semi-SupervisedNetworks}. Using an SAE can be viewed as a form of regularization and prevents the latent representation from over-fitting, which has been shown theoretically to enhance generalizability \citep{Le2018}.  It has also been shown in deep learning that adding auxiliary tasks can act as a form of regularization \citep{Parthasarathy2018LadderAttributes,Liebel2018AuxiliaryLearning,Zhao2019LeveragingCounting}.  Recent work has also developed strategies for unsupervised generation of the tasks \citep{Liu2019Self-SupervisedLearning}.  However, none of these works evaluated how well the inferred latent space fit the generative model and instead focused solely on prediction.

\textbf{Semi-Supervised Learning.} The standard VAE \citep{Kingma2013} (referred to as VAE in the remaining contents) has been widely extended in semi-supervised learning tasks \citep{kingma2014semi, Rezende2015VariationalFlows, joy2021capturing, pmlr-v70-mueller17a} using deep neural networks. These VAE-based semi-supervised models pursue a shared latent space for both inference and prediction, and have shown impressive predictive performance in some applications. However, they have the same limitation that the inferred latent space does not necessarily fit the generative model. In addition, most of existing methods were evaluated only on image data \citep{kingma2014semi, Rezende2015VariationalFlows, joy2021capturing} or sequence data \citep{pmlr-v70-mueller17a}, which might not suitable for our motivating applications (predicting a characteristic, trait or behavior from electrophysiology). Importantly, these models are less interpretable due to the use of complicated generative network structures and thus not suitable for scientific hypothesis generation.

%Meta-learning shares computational techniques.
\textbf{Second-Order Optimization.} Finally, our novel learning technique will require the use of second derivatives to indirectly induce the learned latent variable model to yield a predictive posterior. This is difficult to implement directly via backpropagation. However, by incorporating computational tricks used in some meta-learning and self-supervision techniques \citep{Finn2017Model-agnosticNetworks, Andrychowicz2016LearningDescent, Liu2019Self-SupervisedLearning}, this can be efficiently implemented in modern learning platforms.

%% file: texfiles/methods.tex
\section{Methods}
\label{sec:Methods}

We first introduce notation. 
Let $\{\x_i\}_{i=1,\dots,N}\in\R^p$ be $N$ independent samples with associated outcomes $\{y_i\}_{i=1,\dots,N}\in\mathcal{Y}$ drawn from the true joint probability distribution $p_d(\x,y)$. Given these data, we fit a joint model defined as $\x,\s\sim p_\theta(\x,\s)$ and $y|\s \sim p_\psi(y|\s)$, where $\s\in\R^L$ is the $L$-dimensional latent space. To simplify theoretical derivations we assume that $p_\theta$ and $p_\psi$ are infinitely differentiable, which holds with exponential family models that are commonly used in our applications. We view $\theta$ as parameterizing our generative model while $\psi$ parameterizes the predictive model (the classifier) of the outcome given the latent space.
We assume that a decoder $p_\theta(\x|\s)$ can be derived directly from the likelihood of the true generative model. We approximate our true posterior $p_\theta(\s|\x)$ by a variational encoder $q_\phi(\s|\x)$ to enable variational inference \citep{Kingma2013}. In our model, as in SVAEs or any standard supervised model, the variational approximation is not conditioned on $y$ as this variable is unknown when the model is used for prediction.

%We do not use the posterior $p_{\theta,\psi}(\s|\x,y)$ as $y$ is unknown when the model is used for prediction.

\subsection{Supervised Variational Autoencoders}
\label{ssec:mSVAE}

If our only objective were to obtain a generative model of $\x$, we could simply parameterize $q_\phi$ with a flexible neural network and maximize the Evidence Lower Bound Objective (ELBO) used in VAEs. However, focusing exclusively on the generative model often yields poor predictions of $y$ \cite{Talbot2020SupervisedActivity}. This motivates the inclusion of a supervised loss, forming the SVAE objective
\begin{equation}\label{eq:svae}
\begin{split}
    \calL_{\phi,\theta,\psi}= \E_{p(\x)}\Bigl[&\E_{q_\phi(\s|\x)}\bigl[\log p_\theta(\x,\s) \\ -&\log q_\phi(\s|\x) + \lambda \log p_\psi(y|\s)\bigr]\Bigr], 
\end{split}
\end{equation}
where $\lambda$ functions as a tuning parameter controlling the relative weight of the supervised loss. This objective can be approximated by empirical risk minimization of the observed data and estimated via gradient descent, as is commonly done with VAEs. For simplicity, we omit $\E_{p(\x)}$ in further derivations so our losses are defined for a single data sample. We note that in practice we would use an empirical risk minimization formulation rather than an expectation. If $\lambda=0$, then \eqref{eq:svae} reduces to VAE objective. In practice, $\lambda$ is usually set to a fairly high value to emphasize prediction, as the variance of $y$ is often substantially outweighed by the variance of $\x$ \cite{Talbot2020SupervisedActivity}. 

The SVAE approach may appear ideal: the encoder $q_\phi$ simultaneously provides an accurate reconstruction of $\x$ and a latent representation predictive of $y$.
However, the inclusion of the predictive loss induces a systematic deviation of the encoder from the posterior distribution of the generative model, which we refer to as the deviance. This can be seen by analyzing the stationary point associated with $\phi$ in Proposition \ref{pp1}.
\begin{prop}\label{pp1} The stationary points of \eqref{eq:svae} can be found using the reparameterization  trick used by Kingma \emph{et al.} \cite{Kingma2013}. This reparameterization expresses the random variable $\s\sim q_\phi(\s|\x)$ as a transformation of a random variable $\epsilon$ dependant on the observed data $\x$, denoted $g_\phi(\epsilon,\x)$. Under this transformation, the stationary point such that $\nabla_\phi\calL_{\phi,\theta,\psi}=0$ is
\begin{equation}\label{eq:fp_svae}
\begin{split}
    \E_{p(\epsilon)}\bigl[\nabla_\phi& \log p_\theta(\x,g_\phi(\epsilon,\x)) - \log q_\phi(g_\phi(\epsilon,\x))\bigr] = \\&-\lambda \E_{p(\epsilon)}\bigl[\nabla_\phi\log p_\psi(y|g_\phi(\epsilon,\x)) \bigr].
\end{split}
\end{equation}
The proof and the stationary points for $\theta$ and $\psi$, given $\phi$, match a VAE and are provided in the Supplemental Section \ref{ssec:aFixedPoints}.
\end{prop}

In the case where $\lambda=0$ the left hand side of \eqref{eq:fp_svae} must also be $0$. This represents the standard stationary point of a VAE learned without the supervised loss. In a VAE, the values $\phi$ minimize the divergence of the variational approximation and the true posterior defined by $\theta$. Thus, the right hand side is a systematic deviation from the posterior induced in the latent representation by the predictive objective. This deviation ensures increased relevance of the latent space to the outcome $y$, as larger values of $\lambda$ correspond to an increased emphasis on the predictive objective. As $\theta$ is dependent only on the generative loss, the strength of this deviation indicates the amount of information relevant to $y$ that is not included in the purely generative model.  

\subsection{A Metric of Scientific Utility for Experimental Design}

While scientific goals are highly variable and cannot be completely captured by a single metric, we now explain why this deviation reduces scientific utility in our application. % before introducing a novel metric to quantify this utility. 
 The ultimate goal of our work is to experimentally validate the relationship between the covariates $\x$ and the outcome $y$. The generative model defines this relationship as
\begin{equation}
    p_{\theta,\psi}(y|\x) = \int p_\psi(y|\s)p_\theta(\s|\x)d\s,
\end{equation}
which posits an hypothesis as to how changing $\x$ will impact $y$ based on the parameters $\theta$ and $\psi$.  Using such a model, one can design an experiment to test whether the relationship defined by the model is causal. For example, this approach has been used successfully to design neurostimulation protocols \cite{Carlson2017,Hultman2018Brain-wideVulnerability,Mague2020Brain-wideState}.

Certain applications may require real time estimation of the latent space and outcomes\cite{Carlson2017}. As described previously, a common fast approximation is to use a variational approximation to obtain a predictive distribution as
\begin{equation}
    p_\phi(y|\x) = \int p_\psi(y|\s)q_\phi(\s|\x)d\s.
\end{equation}

Unfortunately, $\theta$ provides no direct insight on $p_\phi(y|\x)$. However, interpreting $\theta$ on how to influence $p_\phi(y|\x)$ is reasonable when $p_{\theta}(y|\x)$ is ``close'' to $p_\phi(y|\x)$, with this approximation degrading in quality as these distributions diverge. This divergence can be quantified via a variety of ways, and we say a model has high scientific utility if this divergence is small. 
%This utility can then be used as a model-selection method in a standard fashion.

Unfortunately, this measure typically cannot be computed exactly without expensive MCMC methods. However, if we modify our definition of scientific utility slightly to use a KL-divergence on the latent space $p(\s|\x)$ \citep{Kingma2013}, we can obtain a reasonable metric that is straightforward to compute. This is reasonable as the chosen relationship between $\s$ and $\y$ is a Lipschitz function, ensuring that small posterior divergences will correspond to small predictive differences. 

This measure can be estimated as follows. We replace the true posterior $p_\theta(\s|\x)$ with the encoder learned by a VAE $q_{\phi^*}(\s|\x)$ holding $\theta$ fixed. By definition, the VAE encoder minimizes the KL from the true posterior out of all variational approximations. By extension, any approximation close to the variational posterior will also have a small KL from the true posterior. As both $q_{\phi^*}(\s|\x)$ and $q_{\phi}(\s|\x)$ are analytic (usually Gaussian), this divergence is straightforward to compute using an analytic formula. 

In summary, we measure scientific utility as the KL between the supervised encoder and a ``refit'' VAE encoder holding the decoder and classifier parameters fixed. Small divergences suggest that the parameters $\theta$ provide reasonable insight on how to modify $\x$ to induce changes in $y$. Furthermore, the systematic discrepancy that SVAEs introduce demonstrated in the previous section inherently reduce this utility.

\subsection{A Bilevel Objective to Maintain a  Posterior Approximation}
\label{ssec:mSOSVAE}

%A VAE has high potential utility in our applications because the stationary point provides the best approximation to the generative posterior by a flexible neural network. However, it only contain much predictive information when the outcome explains much of the data variance, which is uncommon in many of our motivating applications \citep{Talbot2020SupervisedActivity}. An SVAE rectifies the latter issue but induces a deviation in the encoder from the generative posterior.
As described previously, an SVAE learns a system that predicts accurately but can induce significant bias, decreasing utility in our applications.
We mitigate this issue by  ``recoupling'' the encoder of an SVAE to match a VAE encoder via a constraint while still including the predictive loss in the objective function. This combined objective is
\begin{equation}\label{eq:sosvae_1_bilevel}
\resizebox{.91\columnwidth}{!}{$
\begin{aligned}
    \textstyle \max_{\psi,\theta,\phi}\;\; &\E_{q_\phi(\s|\x)}\bigl[\log p_\theta(\x,\s) -\log q_{\phi}(\s|\x)  + \lambda \log p_\psi(y|\s) \bigr]\\
    \textrm{s.t.}\;\; \phi &= \textstyle \argmax_{\phi'} \E_{q_{\phi'}(\s|\x)}\bigl[\log p_\theta(\x,\s)-\log q_{\phi'}(\s|\x)\bigr].
\end{aligned}
$}
\end{equation}
This objective corresponds to a bilevel optimization problem \citep{dempe2015bilevel}. To match the terminology of the bilevel optimization literature, the higher-level objective, which we denote $F(\theta,\psi,\phi)$, matches an SVAE while the lower-level objective, which we denote $f(\phi,\theta)$ constrains the encoder to approximate the posterior. For notational convenience, we define $F_x=\E_{q_\phi(\s|\x)}\bigl[\log p_\theta(\x,\s) -\log q_{\phi}(\s|\x)]$ and $F_y=\E_{q_\phi(\s|\x)}\bigl[\lambda \log p_\psi(y|\s)]$ as they represent the generative and supervised aspects of the higher-level objective respectively.

We assume that this is an opportunistic problem, signifying that the lower-level objective is not actively opposing the primary objective \citep{dempe2007new}. Furthermore, we assume that the lower-level objective is unconstrained, as careful reparameterization of neural network parameters can remove most constraints. Even with these assumptions, bilevel optimization problems are notoriously difficult to analyze \citep{dempe2015bilevel}. One common approach is to reformulate the the bilevel optimization as a standard optimization problem. We can achieve this by replacing the lower-level objective with a standard constraint that the gradient $\nabla_\phi f=0$, corresponding to replacing the lower-level objective with its Karush-Kuhn-Tucker (KKT) conditions \citep{humpherys2021foundations}. These two objectives are equivalent provided that $f$ is convex \citep{dempe2012karush}. This assumption holds over $\phi$ in the simplified case of a supervised autoencoder with an affine encoder, a model that has been used previously in multiple applications \citep{mague2022brain,Talbot2020SupervisedActivity}. However, this is unlikely to hold once this is reformulated as a variational autoencoder, and certainly will not hold once $\phi$ corresponds to a deep neural network. When convexity fails to hold, not only can suboptimal stationary points be discovered, the global optimum may fail to be a stationary point \citep{mirrlees1999theory}.

We will continue with this substitution even when it violates convexity, which we believe is reasonable approach in practice to build intuition. First, recent work has shown that the function $g_\phi$ is convex around the optimum in the function space in the infinite width limit \citep{jacot2018neural,allen2019convergence}. 
The functional form is our primary interest, and the specific parameters of the encoder are irrelevant in our applications. Remarkably, in the finite width regime overparameterized neural networks also possess similar properties to convex problems \citep{ergen2021convex}. Second, even local optima in neural networks yield state-of-the-art performance in many tasks. 
Work has shown that high-energy stationary points are often saddle points \citep{draxler2018essentially}, which can be largely avoided with modern gradient descent techniques \citep{jin2017escape}. Thus, the stationary point associated with this substitution is likely to be reasonably close to the global optimum with models that have been used with great success in a variety of practical tasks.

Ultimately, we rely on a commonly used procedure in machine learning. We first develop an approach using strong assumptions and then provide intuitive justifications for why the approach is reasonable even when the assumptions are not strictly true, and finally we validate these claims empirically with practical applications. 

Having replaced the lower level optimization with a gradient constraint, our objective now is 
\begin{equation}\label{eq:sosvae_1}
\resizebox{.91\columnwidth}{!}{$
\begin{aligned}
    &\textstyle \max_{\psi,\theta,\phi}\;\; \E_{q_\phi(\s|\x)}\bigl[\log p_\theta(\x,\s) -\log q_{\phi}(\s|\x)  + \lambda \log p_\psi(y|\s) \bigr]\\
    &\textrm{s.t.}\;\;  \nabla_\phi\E_{q_{\phi'}(\s|\x)}\bigl[\log p_\theta(\x,\s)-\log q_{\phi'}(\s|\x)\bigr]=0.
\end{aligned}$}
\end{equation}
We now have a single-level optimization problem with a non-convex equality constraint.

\subsection{Derivation of the Learning Algorithm}
We now derive an algorithm for maximizing \eqref{eq:sosvae_1}. We will use stochastic gradient descent to perform inference, as such methods provide computational tractability under the large-data regimes we encounter. However, this descent must be performed moving along the manifold defined by the constraint. The step on $\psi$ is identical to an SVAE, as $\psi$  is not involved in the constraint. The step on $\phi$ is straightforward as well; gradient descent is performed with respect to the constraint until convergence. However, the step for $\theta$ is more challenging.

One intuitive view of the constraint is that it makes $\phi$ a function of $\theta$; the generative parameters determine the best variational approximation to the posterior. If this were a standard function, we could solve for $\phi$ in terms of $\theta$ in the primal objective and bypass the constraint entirely. We know that this is not possible; the non-convexity of the lower-level objective makes the feasible set a  set-valued function. However, we can use the Implicit Function Theorem \citep{humpherys2017foundations} to make this substitution valid within a neighborhood of the current values of $\theta$ and $\psi$.

Let $G:\Theta\times\Phi\rightarrow \mathbb{R}^q$ be defined as the map $G(\theta,\phi)=\nabla_\phi f(\theta,\phi)$ and $q$ is the number of parameters in $\phi$. Our manifold can then be defined as the level set such that $G(\theta,\phi)=\bm{0}$. The Implict Function Theorem states that if the second derivative of $G$ with respect to $\theta$ has a bounded inverse at a point $(\phi_0,\theta_0)$, then there exists an open neighborhood $\Theta_0\times\Phi_0$ wherein an unique differentiable function $h:\Theta\rightarrow\Phi$ exists such that $G(\theta,h(\theta))=G(\theta,\phi)$. This unconstrained objective is
\begin{equation}\label{eq:solved}
\resizebox{.91\columnwidth}{!}{$
    \textstyle \max_{\psi,\theta}\;\; \E_{q_{h(\theta)}(\s|\x)}\bigl[\log p_\theta(\x,\s) -\log q_{h(\theta)}(\s|\x)  + \lambda \log p_\psi(y|\s) \bigr].\\
$}
\end{equation}

We can now obtain the gradients of $\theta$ with respect to this unconstrained objective. The total derivative is 
\begin{equation}
    \frac{dF}{d\theta} = \frac{\partial F_x}{\partial \theta } + \frac{\partial F_x}{\partial h}\cdot \frac{\partial h}{\partial \theta} + \frac{\partial F_y}{\partial h}\cdot \frac{\partial h}{\partial\theta}
\end{equation}
This expression can be simplified by noting that as $\phi$ is learned to convergence with respect to the constraint, we can make the simplifying assumption $\partial F_x/\partial h=0$. Thus, our update on $\theta$ consists of a standard partial derivative with respect to the generative likelihood as well as a second-order term induced by the constraint.

The gradients for $\nabla_\phi F$, $\nabla_\psi F$, $\partial F_x/\partial \theta$ and $\partial F_y/\partial h$ are straightforward, however $\partial h/\partial \theta$ is not. While the Implicit Function Theorem provides an explicit means to evaluate this gradient, it involves computing the Hessian matrix. While this is possible in low dimensions, it is clearly infeasible when modern neural networks containing millions of parameters are used for the encoder. 

Instead, we must find a means to approximate this second-order term. This can be done using an optimization trick used in meta-learning and self-supervised learning \citep{Finn2017Model-agnosticNetworks}. While our work differs dramatically in objective, we share a common inference technique. This approximation replaces the Hessian $\nabla_\theta\nabla_\phi$ with a discrete approximation $\nabla_\theta \delta_\phi$. This can be efficiently implemented in modern packages by maintaining the computational graph but estimating the gradients of $\theta$ using the updated parameters $\phi'$ estimated after learning to convergence. 

In practice, we only use a single update of $\phi$. As the steps on $\theta$ are small, this step $\phi$ maintains the constraint within a small tolerance. The parameters $\psi$ are also updated with a single step, and $\theta$ is updated with a standard gradient descent step along with a second-order step to account for the dependence of $\phi$ on $\theta$. The procedure is summarized in Algorithm \ref{alg:sosvae}.

\SetKwInput{kwInit}{Initialize}
\begin{algorithm}[t]
\SetAlgoLined
\KwIn{$X=\{\x_1,...,\x_N\}\in\mathbb{R}^p$, $Y=\{y_1,...,y_N\}\in\mathcal{Y}$.}
\kwInit{Network parameters: $\phi,\theta,\psi$; learning rate: $\alpha, \beta$; weights: $\lambda.$}
\For{epoch in iterations}{
$(\x_i,y_i), i\in\{1,...,N\}\;\;\;\;\;\;\;\;\;$    \# Batch data\\
$\eta\sim$$p_\theta(s)$, $\epsilon_i\sim$$N(0,I)$ \# Latent space definition\\
% $\eta\sim$$N$$(0,I)$, $\epsilon_i\sim$$g_\phi$$(\epsilon,\x_i)$ \# Latent space definition\\
\# Step 1: decoder update\\
$\theta^+\gets\theta+\alpha\nabla_\theta(\log p_\theta(\x_i,g_\phi(\epsilon_i,\x_i)) - KL(\epsilon_i,\eta))$\\
\# Step 2: encoder update\\
$\phi^+\gets\phi+\alpha\nabla_\phi(\log p_\theta(\x_i,g_\phi(\epsilon_i,\x_i))- KL(\epsilon_i,\eta))$\\
 \# Step 3: classifier update\\
$\psi^+\gets\psi+\alpha\nabla_\psi(\lambda \log p_\psi(y_i|g_{\phi^+}(\epsilon_i,\x_i)))$ \\
\# Step 4: second-order decoder update\\
$\theta^{++}\gets\theta^++\beta\nabla_{\theta^+}(\lambda \log p_{\psi^+}(y_i|g_{\phi^+}(\epsilon_i,\x_i))))$ \\
\# Step 5: Model parameters update\\
    $\psi\leftarrow \psi^+$, $\phi\leftarrow \phi^+$, $\theta\leftarrow \theta^{++}$ 
    }
%     \#Second-order derivative on encoder\;
%     $\phi^'$$\gets\phi^+-\beta\nabla_\phi_$$(\lossy(y_i,h_\psi(f_\phi_^+$$(x_i))))$\;
\caption{\label{alg:sosvae} Second-Order Supervision VAE (SOS-VAE)}
\end{algorithm}

%This formulation yields an unbiased variational approximation of the posterior. This can be seen as $\phi$ is learned via a constraint exclusively on the generative loss. Supervision of the latent space instead is induced in the generative model indirectly via the decoder. This is demonstrated via an analysis of the stationary points in Proposition \ref{pp2}, with the proof deferred to Supplemental Section \ref{ssec:astationaryPoints}.

\subsection{Stationary Point Analysis}

A natural question at this point is how the constraint on $\phi$ is incorporated, as each step in Algorithm \ref{alg:sosvae} is unconstrained. We rely in the strong regularity properties of neural networks for this justification. At each step, the update on $\theta$ moves the current parameter values to a point in $\Theta\times\Phi\times\Psi$ that violates the constraint. However, our update on $\phi$ to convergence returns the parameter values to the manifold defined by the constraint. While in practice this single update approximation has worked well, if necessary $\phi$ can be updated many times to more closely satisfy the constraint.  
%While we will not formally prove convergence, we note that work showing similar optimization procedures do converge in other applications \citep{fallah2020convergence}, and our procedure is empirically successful. 

% Consider readding the next two paragraphs, but they just seem handwavy.  Do we need?

We now sketch a justification for convergence of our algorithm.  We first note that the discrete approximation trick used in our algorithm has been shown to converge under mild regularity conditions in meta-learning \citep{fallah2020convergence}.
%While some steps lack the rigor to constitute a formal proof, work showing that similar optimization procedures converge in other applications lend credence to our claims \citep{fallah2020convergence}. 
Based on the  smoothness assumptions of the objective, we can ensure that the primal objective increases at each iteration by selecting a sufficiently small step size. As our objective is a constrained variational autoencoder, it automatically inherits the property of being a lower bound on the marginal log likelihood. This likelihood often corresponds to a simple generative model and consequently we can assume this likelihood is finite. As a bounded monotonically increasing sequence, the primal objective is guaranteed to converge. Furthermore, local convexity in the function space ensures that the model parameters, $\theta$ and $\phi$, will be a local optimum. In summary, our algorithm will converge given a set of assumptions that have held empirically in a variety of applications, namely convexity in the function space and smoothness of the networks.

%Provided these assumptions hold, the previous argument guarantees convergence in the function space. Convergence in the parameter space is reasonable due to the local convexity properties of neural networks previously cited. The guarantees using gradient descent on the risk also extend to stochastic gradient descent on the empirical risk, with the proofs contained in \citet{Bottou2018}. 

%Having provided arguments for the convergence of our algorithm, 
Assuming our given algorithm converges, we can now analyze properties of the resulting stationary point. In particular, we now provide a necessary condition the resulting stationary points. The procedure is largely identical to the stationary points for an SVAE, and hence are deferred to the supplementary material. The only difference is the second-order term stemming from the constraint must be considered with respect to $\theta$.  

\begin{prop}\label{pp2} The stationary points of \eqref{eq:sosvae_1} can be found using the same  reparameterization trick in Proposition \ref{pp1}. The stationary point of $\phi$ matches a VAE and is
\begin{equation}\label{eq:fp_sosvae_phi}
\E_{p(\epsilon)}\bigl[\nabla_\phi \log p_\theta(x,g_\phi(\epsilon,x)) - \log q_\phi(g_\phi(\epsilon,x))\bigr]=0.
\end{equation}
However, $\theta$ is modified to induce a predictive posterior and has a stationary point of
\begin{equation}\label{eq:fp_sosvae}
\resizebox{\columnwidth}{!}{
    $
    \E_{p(\epsilon)}\bigl[\nabla_\theta \log p_\theta(x,g_\phi(\epsilon,x))\bigr] = -\lambda \E_{p(\epsilon)}\bigl[\nabla_\theta\log p_\psi(y|g_\phi(\epsilon,x)) \bigr].
    $}
\end{equation}
The stationary point form for $\psi$ given $\phi$ matches an SVAE.
\end{prop}

\begin{figure*}[t]
    \centering
    \includegraphics[width=0.85\textwidth]{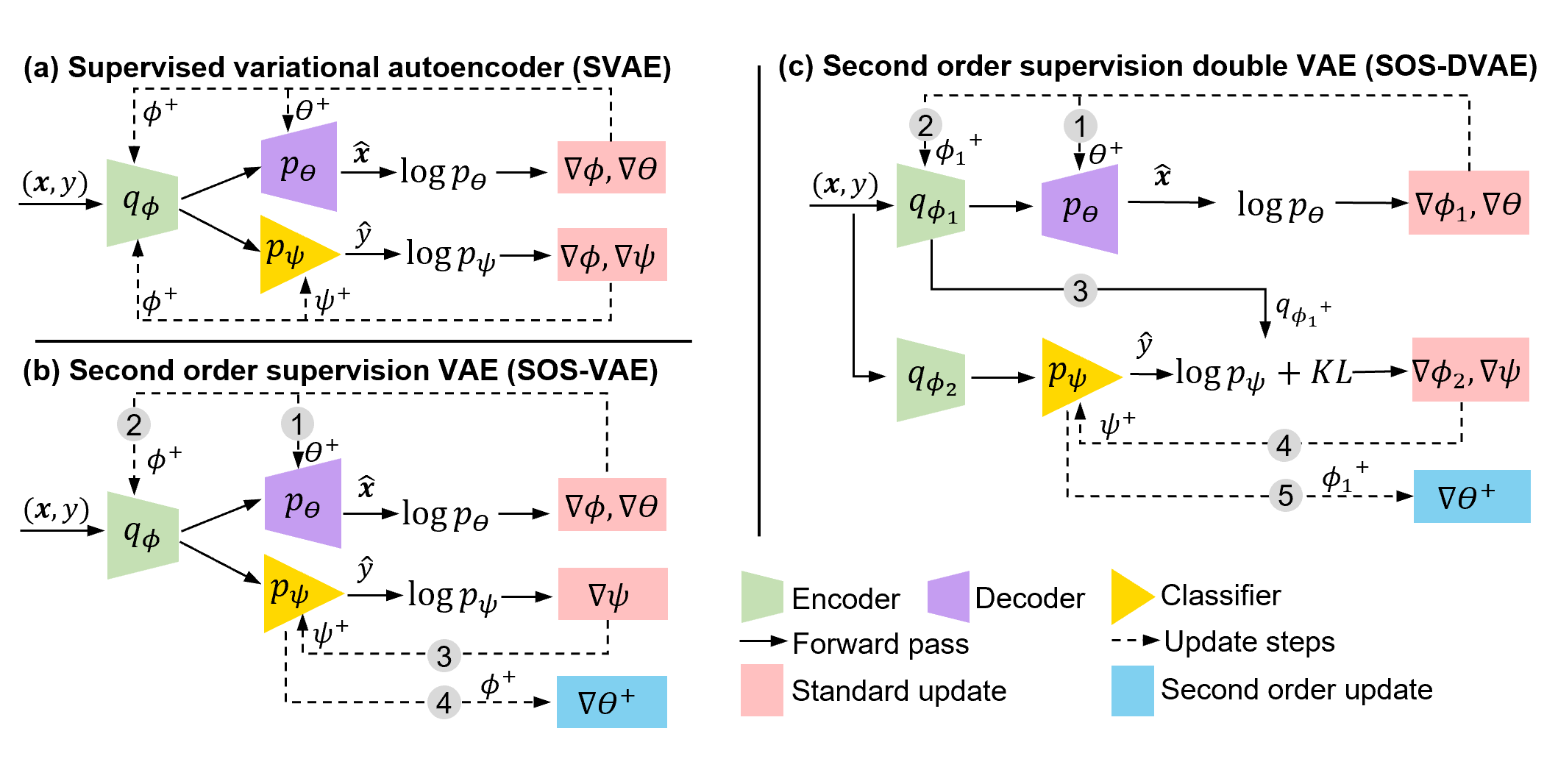}
    \vspace{-3mm}
    \caption{Visualizing the learning procedures of (a) SVAE (b) SOS-VAE and (c) SOS-DVAE. The losses and parameters are defined in the text. $\bm{x}$ and $y$ denote an arbitrary sample. In (b), we mark the 4 learning steps from Algorithm \ref{alg:sosvae}.  The major difference is that a second-order step is used on the supervised loss to update the generative model parameters rather than the encoder.  In (c), the 5 update steps from Supplemental Algorithm \ref{alg:sosdvae}. The major change from SOS-VAE is that a second encoder is used to relax the exact inference strategy and measure divergence.}
    \label{fig:overview}
\end{figure*}

The stationary points of the SOS-VAE provide an interesting contrast with an SVAE. Our modified objective induces a predictive latent space by influencing the generative parameters, hence our colloquial description of supervising the decoder. As the stationary point of the encoder is solely with respect to the generative model, refitting an encoder with the VAE loss and a fixed $\theta$ will not meaningfully change the encoder, making sure we maintain utility for our applications.  We demonstrate this claim empirically in Section \ref{ssec:MNIST}.

%% file: texfiles/extensions.tex
\section{Relaxing the SOS-VAE Framework}
\label{sec:Extensions}

SVAEs can be viewed as a regularization technique on a flexible classifier \citep{Le2018}. In contrast, the SOS-VAE explicitly uses the generative model to induce a predictive latent space. If a need for predictive accuracy matches or exceeds the need for informative decoder parameters, it may be preferable to allow the encoder to deviate slightly from the posterior to obtain increased flexibility, as a slight relaxation can yield substantial gains in predictive accuracy with only a minimal decrease in scientific utility. This improved performance can be incredibly valuable when the model is used to track the auxiliary variable, such as in real-time stimulation experiments \citep{armstrong2013closed}.
We thus extend SOS-VAE by relaxing the constraint on the encoder, which we refer to as the Second-Order Supervision Double VAE (SOS-DVAE). We show that this approach can also be used to model missing data in Appendix \ref{ssec:eMissingData}. Such missingness commonly arises when datasets obtained from multiple experiments are combined to broaden conclusions and increase statistical power or do to common electrode failure. 

% \subsection{Relaxing the Encoder Constraint}
\label{ssec:eRelaxing}

The SOS-DVAE approach defines two distinct encoders, a generative encoder $q_{\phi_1}(\s|\x)$ that exclusively approximates the generative posterior and a predictive encoder $q_{\phi_2}(\s|\x)$ that includes the supervision task. The generative encoder is learned as in Section \ref{ssec:mSOSVAE} to approximate the true generative posterior. However, the predictive encoder is learned to maintain high predictive ability but with regularization towards the generative encoder. One objective that allows for this tradeoff is
\begin{equation}\label{eq:sosvae}
\begin{split}
    \textstyle \max_{\psi,\theta,\phi_2}\;\; &\textstyle \E_{q_{\phi_2}(\s|\x)}\bigl[\log p_\theta(\x,\s)+\lambda \log p_\psi(y|\s)\bigr] 
     \\&\;\;\;\;\;+\E_{q_{\phi_2}(\s|\x)} \bigl[\lambda \log p_\psi(y|\s)\bigr]- \mu KL(q_{\phi_1},q_{\phi_2})\\
    \textstyle \textrm{s.t.}\;\textstyle \phi_1 =& \argmax_{\phi'} \E_{q_{\phi'}(\s|\x)}\bigl[\log p_\theta(\x,\s)-\log q_{\phi'}(\s|\x)\bigr].
\end{split}    
\end{equation}
Here, $\mu$ is a tuning parameter controlling the strength of regularizing $\phi_2$ to yield proper posterior inference (as $\mu\rightarrow\infty$, $\phi_2$ is forced to exactly match $\phi_1$). The parameters of the generative encoder, $\phi_1$, are learned in the same way as $\phi$ in \eqref{eq:sosvae_1}, meaning that it approximates the posterior of the generative model while inducing the generative model parameters to be useful for prediction. The parameters of the predictive encoder $\phi_2$ are free to deviate in approximating the generative posterior in order to yield improved predictions, provided that $q_{\phi_2}$ is close to the generative encoder as measured by a KL regularization term. The KL term also allows straightforward estimation of how much deviation there is from the generative model. We note that alternative distribution distances could be used if desired, and KL is one approach that provides straightforward gradients. This learning approach is visualized in Fig. \ref{fig:overview}c and follows the same logic as SOS-VAE (Fig. \ref{fig:overview}b). 

Once again, we proceed using by replacing the constraint with its KKT conditions and rely on the previously-provided justifications for this substitution. Pseudo-code is provided in Supplemental Algorithm \ref{alg:sosdvae}. We note that removing the second-order optimization step is empirically detrimental to the predictive accuracy and increases the deviation of the variational approximation. This importance of the second-order step contrasts with other work that uses this technique, which found that such a change was largely inconsequential \citep{Finn2017Model-agnosticNetworks}.

%% file: texfiles/results.tex
\section{Experimental Results}
\label{sec:results}

We present empirical evaluations on the SOS-VAE and SOS-DVAE inference strategies compared to several baseline approaches: \textbf{1}) \textit{VAE-refit} \textemdash a sequential fitting strategy with a VAE approximating a generative model and then using a supervised network on the frozen latent representation (this matches the ``cutting-the-feedback'' strategy in statistics \citep{Plummer2014}), \textbf{2}) \textit{SVAE} \textemdash the standard SVAE technique, \textbf{3}) \textit{SVAE-refit} \textemdash an SVAE where the encoder is refit to optimize only the generative objective while the supervised component remains fixed after initial training, and \textbf{4}) \textit{SDVAE} \textemdash the SOS-DVAE approach without the second-order step (Supplemental Algorithm \ref{alg:sdvae}). These strategies are chosen to compare different aspects of scientific utility. The VAE-refit enforces that the encoder approximate the generative posterior distribution as well as possible, at the cost of a potentially less predictive latent space. The SVAE will yield good predictive and reconstructive performance but has a biased approximation of the posterior distribution. Comparing the performance of SVAE to that of SVAE-refit reveals this inherent bias, as predictive improvements often do not induce changes in the generative parameters. Finally, SDVAE illustrates the need for our second-order strategies. 

We first demonstrate our methods on the MNIST dataset as a proof-of-concept to illustrate the issues associated with a biased approximation. We then demonstrate scientific applications on two neural recording datasets, one consisting of LFPs recorded in mice and the other consisting of Electroencephalography (EEG) measurements in humans.

 In each application, we use KL to evaluate the discrepancy between the posteriors from the two encoders in the SOS-DVAE caused by the supervision. There are other commonly-used methods for comparing distribution similarity \citep{basu1998robust,van2014renyi,muller1997integral}. However, KL is chosen for two important reasons. First, the likelihood is nearly ubiquitous in statistics to evaluate model fit, and as this work is motivated by generative modeling the KL is a natural choice. More importantly, these models are learned to minimize this KL, and evaluating similarity using the KL makes it clear that the bias stems from the supervised objective rather than the similarity metric used for training.

\begin{figure}[ht]
\centering
\includegraphics[width=0.99\columnwidth]{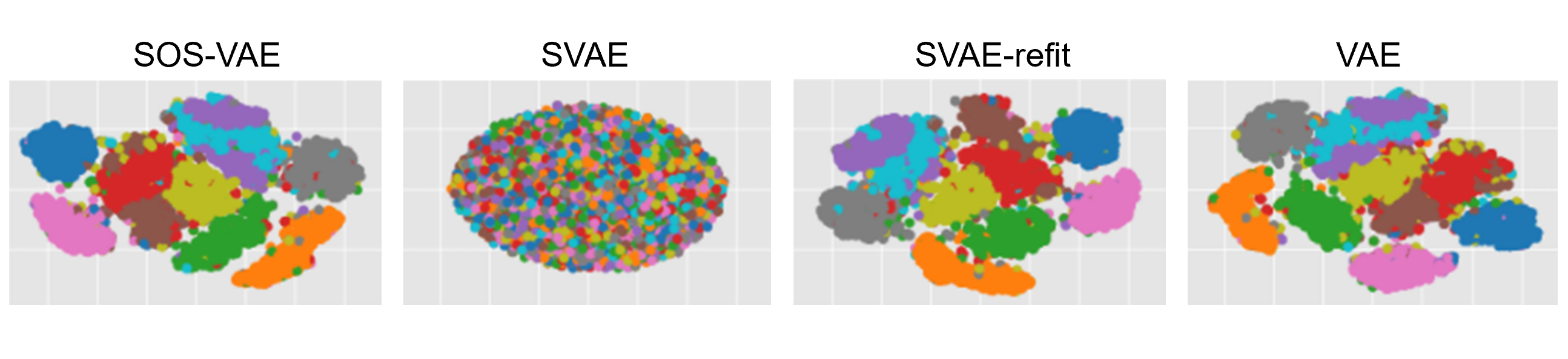}
\caption{T-SNE visualization of the learned latent representation space for MNIST dataset using each model type. Each class is represented by a different color. $\lambda=1e4$, feature size of 28x28, learning rate of 1e-3, and a batch size of 128 were used in the SVAE model.}
\label{fig:latent_tsne}
\end{figure}

A Multilayer Perceptron (MLP) with a single hidden layer is used as the encoder for all experiments.  We evaluate two types of generative decoders.  First, we use a Non-negative Matrix Factorization (NMF), which has been frequently used as a model capable of uncovering latent networks from neural electrophysiological data \citep{Gallagher2017Cross-spectralAnalysis,Talbot2020SupervisedActivity,khambhati2018modeling,Medaglia2015}. Second, we use an MLP decoder with a single hidden layer to demonstrate the broad applicability of the methods. All models have a latent space dimension of 20. These parameter choices are made to match common settings where supervised generative models have been used \citep{mague2020brain,block2020prenatal}. Additional details on implementation and training are given in the Supplemental Section \ref{sup:implementationalt}. The code to reproduce these result is available on Github\footnote{https://github.com/liyuntu/SOS-DVAE}.

\subsection{Demonstrating the Effect of Posterior Bias}
\label{ssec:MNIST}

To demonstrate the consequences of bias in the variational approximation of a standard SVAE and the effectiveness of our proposed solution, we perform baseline comparisons in the familiar MNIST database of handwritten digits~\citep{deng2012mnist}. 
The model is trained and evaluated using 10-fold cross-validation on the  70,000 images.  Full training parameters are in Supplemental Table~\ref{st:params}. 

\begin{figure}[t]
\centering
\includegraphics[width=\columnwidth]{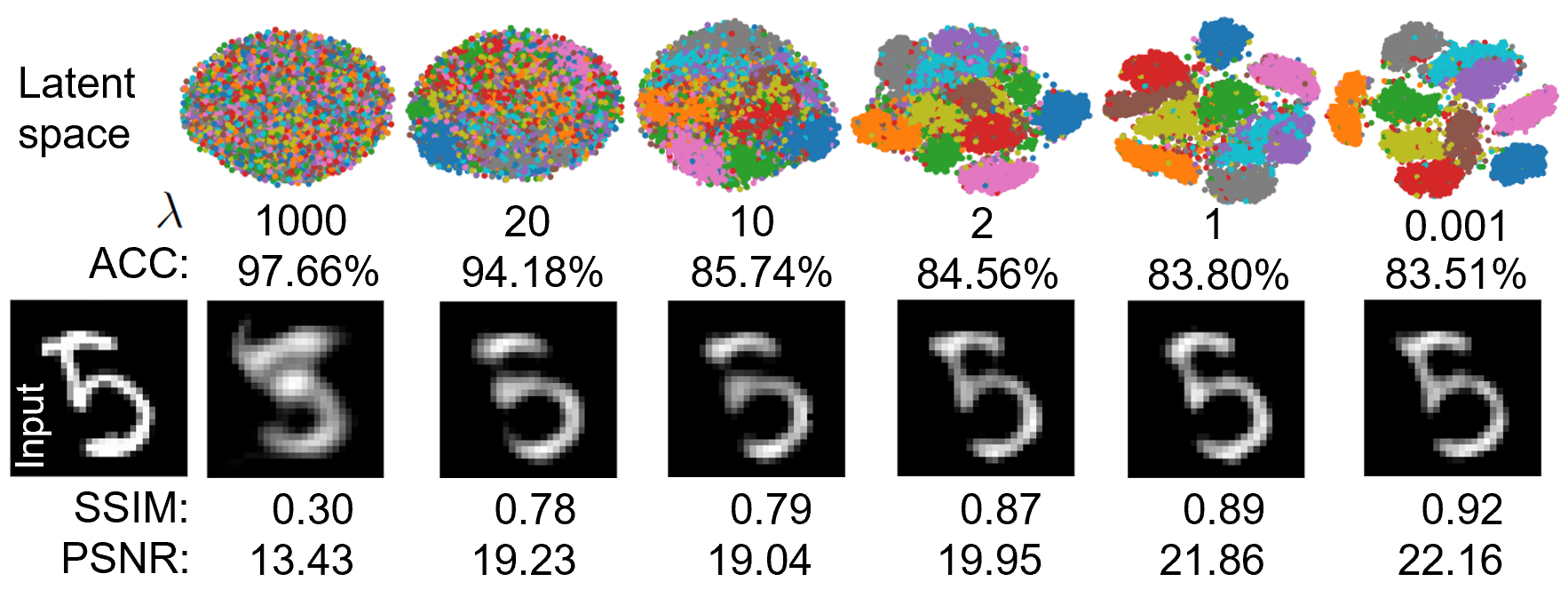}
\caption{Prediction and generation performance of an SVAE model on the MNIST dataset using MLP encoder as a function of $\lambda$. The top row shows a T-SNE embedding of the latent representations. ACC = Test set accuracy; SSIM = Structural Similarity Index; PSNR = Peak Signal-to-Noise Ratio. Bigger values of SSIM and PSNR mean the generated image is closer to the input image. The leftmost image is a representative example input to the model; other images are representative of what is generated by the corresponding model for that input. It use the same hyperparameters as Fig. 3.}
\label{fig:tune_lambda_svae_mnist_mlp}
\end{figure}

We first visualize the effect of the bias in SVAE models on the distribution of the latent representations in Fig. \ref{fig:latent_tsne}.
We find that the latent space of the SVAE model is changed to the point that there is no resemblance to the latent space of the fully generative VAE model.
In contrast, the latent space for the SOS-VAE and SVAE-refit models closely resemble that of the VAE.
The massive change in latent space from SVAE to SVAE-refit clearly explains why predictive performance is lost in the refit model
Fig. \ref{fig:tune_lambda_svae_mnist_mlp} shows the effect of tuning $\lambda$ for an SVAE.
We use Peak Signal-to-Noise Ratio (PSNR) and Structural Similarity Index (SSIM) to measure the quality of the generated images relative to the original.
When $\lambda$ goes higher, the generated images are of lower quality when compared to the original. Along with visualizations of the latent space, this demonstrates that the generative structure in the latent space is lost in exchange for good predictive accuracy with high $\lambda$.

\begin{table*}[h]\centering
    \caption{Prediction performance on MNIST and SEED dataset with NMF and MLP decoders.}
    \vspace{3mm}
%    \resizebox{\columnwidth}{!}{
\resizebox{\textwidth}{!}{
    \begin{tabular}{lcccccccccccc}
    \hline\hline
    &\multicolumn{5}{c}{MNIST} & & \multicolumn{5}{c}{SEED}\\
    \cline{2-6} \cline{8-12}
        &\multicolumn{2}{c}{MLP} & & \multicolumn{2}{c}{NMF} &&\multicolumn{2}{c}{MLP} & & \multicolumn{2}{c}{NMF} \\
        \cline{2-3} \cline{5-6} \cline{8-9} \cline{11-12}
        & ACC (\%) & AUC && ACC (\%) & AUC && ACC (\%) & AUC && ACC (\%) & AUC\\
         \hline
         VAE-refit & $89.33\pm0.88$& $0.94\pm0.005$ && $63.46\pm1.23$ & $0.79\pm0.007$ && $46.56\pm5.87$ & $0.60\pm0.043 $&& $35.61\pm0.74$ & $0.52\pm0.005$\\
        SVAE & $97.66\pm0.22$ & $0.99\pm0.001$ & & $97.12\pm0.22$ & $0.98\pm0.001$ & & $61.05\pm5.76$ & $0.71\pm0.043$ & & $60.19\pm7.38$ & $0.70\pm0.054$\\
        SVAE-refit & $15.97\pm4.09$ & $0.53\pm0.023$ & & $65.66\pm1.21$ & $0.81\pm0.007$ & & $35.85\pm2.58$ & $0.52\pm0.021$ & & $35.33\pm0.54$ & $0.52\pm0.005$\\
        SOS-VAE  & $93.08\pm0.40$ & $0.96\pm0.002$ & & $65.24\pm1.02$ & $0.80\pm0.006$ & & $52.60\pm7.02$ & $0.64\pm0.052$  & & $37.11\pm2.25$ & $0.53\pm0.017$ \\
        SDVAE & $98.31\pm0.17$ & $0.99\pm0.001$ & & $97.74\pm0.29$ & $0.99\pm0.002$  & & $60.43\pm6.21$ & $0.70\pm0.046$ && $59.98\pm5.97$ & $0.70\pm0.044$\\
        SOS-DVAE & $98.30\pm0.19$ & $0.99\pm0.001$ & & $97.77\pm0.21$ & $0.99\pm0.001$ & & $60.44\pm6.26$ & $0.70\pm0.046$ &&  $60.04\pm6.14$ & $0.70\pm0.045$\\
    \hline
    \end{tabular}
%    }
}
\label{table:predictionperformance}
\end{table*}

Next, we evaluate predictive performance of our methods compared to the baselines previously mentioned.
Predictive performance is quantified via accuracy (ACC) and area under the ROC curve (AUC) ( Table \ref{table:predictionperformance}). Because MNIST is a multi-class classification problem, the AUC is averaged over all individual classification tasks.
%From these results there are several critical observations we can make. 
Unsurprisingly, among the models that use an MLP decoder there is little variation between our novel inference method and the predictive baselines. The ``cutting-the-feedback'' method had lower accuracy due to the latent space not focusing on prediction, but the classes are still largely distinguishable in this latent representation. However, refitting the encoder of the SVAE (the refit encoder approximates the true posterior of the SVAE generative model) resulted in a dramatic drop in predictive performance. In other words, the bias introduced when estimating the encoder is absolutely critical for maintaining predictive ability and this predictive information is not incorporated into the unbiased generative model. 

%The generative parameters learned by the SVAE would yield highly misleading conclusions if they were used for scientific exploration in a manner similar to our applications.

The results of the models using an NMF decoder are equally interesting, yielding patterns that are dramatically different from the MLP decoder. Here, the SOS-VAE results in a dramatic drop in predictive ability relative to the SVAE. This drop is unsurprising due to the restrictive constraint imposed by shallow generative models that was noted previously \citep{Hahn2013}. As an NMF decoder is substantially less complex than an MLP, we would expect this constraint to become more apparent. However, the SOS-DVAE, by relaxing this stringent condition, is able to maintain a high predictive accuracy marginally superior to an SVAE. The novel method without the second-order step (SDVAE) is also able to maintain high predictive accuracy.

\begin{figure*}[tb]
    \centering
    \includegraphics[width=0.65\textwidth]{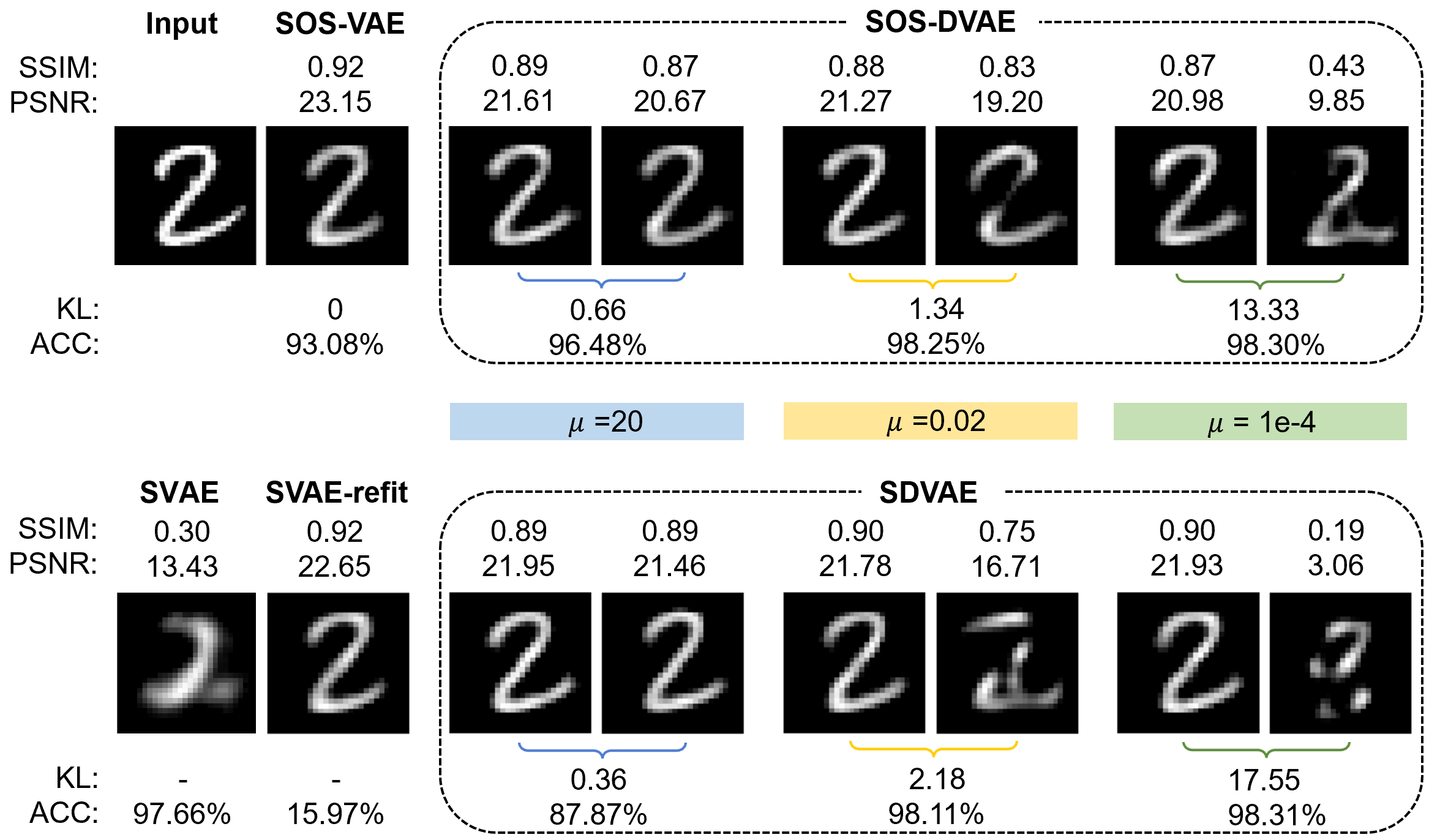}
    \caption{Comparison of predictive and generative performance using an MLP decoder on MNIST. The prediction accuracy is reported below the example reconstruction from the generative portion of each model. In the bottom left, we show the results for a model learned as an SVAE before and after (SVAE-refit) updating the encoder to prioritize the generative model. The results of SOS-DVAE and SDVAE are shown for 3 different KL in nats between the posteriors given by $\phi_1$ and $\phi_2$ (controlled by $\mu$, see Section \ref{ssec:eRelaxing}). For each $\mu$, the left and right pictures are reconstructed from the generative encoder $f_{\phi_1}(\cdot)$ and the classification $f_{\phi_2}(\cdot)$ encoder, respectively. The PSNR and SSIM are both calculated as the average over 2000 randomly selected samples from the test set between the input and reconstructed samples. We note that the SSIM and PSNR for VAE are 0.92 and 22.94 respectively, which are omitted here due to limited space. Higher values are desirable for both metrics. $\lambda=1e4$ was used for each model. Bigger values of SSIM and PSNR mean the generated image is closer to the input image.}
    \label{fig:mnist_generative}
\end{figure*}

We now analyze the other aspect required for scientific utility, that the latent space estimated by the encoder approximates the posterior of the generative model. While it is difficult to inspect this posterior, we can visualize the reconstruction of samples to determine the quality of the variational approximation to the posterior defined by the decoder. An example of this is shown in Fig. \ref{fig:mnist_generative} using an MLP decoder. The top left represents the original image, with the remaining images depicting the reconstructions provided by various models. The SOS-VAE strongly resembles the original image, which is unsurprising as it is an unbiased approximation to the true posterior. Below, we show the reconstruction of the SVAE, along with the reconstruction after the encoder is refit. Both methods provide reasonable reconstructions of the image, but the refit SVAE has substantially worse predictive performance. This is precisely why bias in the encoder has remained undetected; the SVAE provides reasonable reconstructions and excellent predictions. It is only when the generative parameters are used, either for causal manipulations or scientific interpretation, that the bias manifests itself.

Fig. \ref{fig:mnist_generative} visualizes that the results of the SOS-DVAE as we increasingly relax the requirement that the variational approximation remains unbiased (smaller values of $\mu$). Each pair represents models learned for a particular strength, with the left representing the reconstruction of the generative encoder $f_{\phi_1}$ while the right represents the reconstruction using the predictive encoder $f_{\phi_2}$. We can see that a strong emphasis on an unbiased encoder result in minor gains in predictive accuracy, small divergences between the two encoders, and similar reconstructions from both encoders. However, the predictive encoder $f_{\phi_2}$ increasingly diverges from the generative encoder as this regularization shrinks, yielding increasingly poor reconstructions from the predictive encoder. However, this relaxation in alignment between the two encoders corresponds to increasingly improved predictions, emphasizing the utility of $\mu$ as a tuning parameter. In particular, it is worth noting that a substantial improvement in predictive ability can be obtained with minimal increase in bias as shown when $\mu=20$.

Finally, we can analyze the impact of the second-order update by performing a similar analysis without this step (SDVAE), with results shown in the bottom row. We can see that the predictive accuracy is worse almost uniformly over the entire range. Furthermore, while the bias in the encoder is smaller relative to the SOS-DVAE initially, it quickly becomes substantially larger. This results in almost unrecognizable reconstructions using the predictive encoder at the weakest regularization strengths. From this, we conclude that the novel second-order optimization step developed in this work substantially improves both predictive and generative performance.

\subsection{Comparison with Other Semi-supervised Generative Models}
\label{subsec:comparision_with_others_dgm}

We are also interested in how realistic the reconstructions provided by our unbiased and relaxed approximations are as compared to alternative deep learning techniques with much more complex generative networks. This is evaluated on the MNIST dataset for the same reason as stated above; unlike LFPs, MNIST is easy to visualize and intuitively evaluate. While there are metrics for evaluating the quality of images, such as the Frechet Inception Distance (FID) \citep{heusel2017gans}, these methods have been developed for a substantially different task and are not reliable for MNIST. Instead, we visually evaluate quality, an approach commonly used to compare generative adversarial networks \cite{Goodfellow2014GenerativeNets}.

We provide a visualization of several MNIST reconstructions in Supplemental Fig. \ref{fig:generative_compare}. We compare the reconstructions using a standard VAE \citep{Kingma2013}, its semi-supervised extension \citep{kingma2014semi}, variational flows \citep{Rezende2015VariationalFlows}, auxiliary deep generative models \citep{maaloe2016auxiliary}, semi-supervised deep generative models with smooth-ELBO \citep{Feng_Kong_Chen_Zhang_Zhu_Chen_2021}, and our SOS-VAE and SOS-DVAE. Our methods clearly provide more realistic reconstructions. The other four models yield blurry and diffuse reconstructions to the point that some reconstructions are illegible. Our methods provided reconstructions with distinct lines and no visibly missing sections. It is particularly interesting that our methods provided sharper-looking reconstructions as compared to the VAE with an exclusively generative focus. However, this development is not necessarily surprising, given the extensive work showing that auxiliary tasks can improve performance as relevant regularization \citep{caruana1997multitask}. 

Together, the above analyses strongly support multiple conclusions. First, the encoder bias identified in Section \ref{sec:Methods} is a serious concern that commonly affects supervised generative models. Second, our developed inference methodology addresses this bias, either completely removing it with the SOS-VAE or a controlled reduction via the SOS-DVAE. And third, the reconstructions provided by our methods are likely more realistic and a better model of the generative dynamics.

\subsection{Modeling Functional Brain Networks}

\begin{table*}[h]\centering
    \caption{Prediction performance on LFP dataset.\\}
    %\resizebox{\columnwidth}{!}{
    \begin{tabular}{lccccccc}
    \hline\hline
        &\multicolumn{3}{c}{MLP} & & \multicolumn{3}{c}{NMF} \\
        \cline{2-4} \cline{6-8}
        & ACC (\%) & AUC & Scientific Utility & & ACC (\%) & AUC & Scientific Utility \\
        \hline
        VAE-refit & $58.27\pm1.01$ & $0.61\pm0.009$ &\textemdash & & $54.76\pm0.80$ & $0.56\pm0.009 $&\textemdash \\
        SVAE & $88.80\pm1.21$ & $0.90\pm0.010$ &$415.06\pm533.45$ & & $88.73\pm1.25$ & $0.90\pm0.010 $&$195.75\pm171.74$ \\
        SVAE-refit  & $35.70\pm1.70$ & $0.52\pm0.013$ &\textemdash & & $44.50\pm0.58$ & $0.51\pm0.004$ &\textemdash \\
        SOS-VAE & $73.80\pm2.22$ & $0.78\pm0.018$ &$1.77\pm0.55$ & & $69.37\pm1.86$ & $0.73\pm0.013$ &$23.37\pm14.67$ \\
        SDVAE & $88.42\pm1.41$ & $0.90\pm0.010$ &$55.70\pm46.28$ & & $88.60\pm1.22$ & $0.90\pm0.009$ &$154.81\pm110.05$ \\
        SOS-DVAE & $88.50\pm1.45$ & $0.91\pm0.010$ &$105.23\pm79.12$ & & $88.64\pm1.34$ & $0.91\pm0.009$ &$141.02\pm126.94$ \\
    \hline
    \end{tabular}
    %}
     \label{table:lfp-acc-auc}
\end{table*}

\subsubsection{Decoding Emotional Affect from Electroencephalography Recordings}

We then apply the inference methods described above to the publicly available \emph{SEED} electroencephalography (EEG) dataset \citep{duan2013differential,zheng2015investigating}.
This dataset includes 15 subjects recorded while watching movie clips designated with a negative/neutral/positive emotion label.
We analyze signals from a curated subset of 19 of the original 62 electrodes (see Supplemental Fig. \ref{fig:EEG_electrodes}).
We split the signals into non-overlapping 1 \emph{s} time windows.
For each window, we calculate the spectral power and coherence for 5 different frequency bands: 1--4 \emph{Hz}, 5--8 \emph{Hz}, 9--12 \emph{Hz}, 13--30 \emph{Hz}, and 31--50 \emph{Hz}, corresponding to the Delta, Theta, Alpha, Beta, and Low Gamma bands typically used in EEG analysis, respectively.
Models are trained to reconstruct the power and coherence values and to classify the emotion label associated with each window.
We use a leave-one-participant-out cross-validation scheme to select hyperparameters~\citep{wong_performance_2015}.

Table \ref{table:predictionperformance} reports the performance of each inferred model.
We see that with the MLP decoder, the SOS-VAE gives much better predictive performance than the SVAE-refit model, indicating that the predictive performance of the SVAE model is driven by features in the latent space that are not relevant to the generative model.
The SOS-DVAE sacrifices relatively little performance compared to the SVAE while we again find that SOS-DVAE can get higher predictive performance at smaller KLs compared to SDVAE (see Supplemental Figs. \ref{fig:kl_curves}c and \ref{fig:kl_curves}d).

\begin{figure*}[t]
\centering
\begin{subfigure}{0.32\textwidth}
\includegraphics[width=\textwidth]{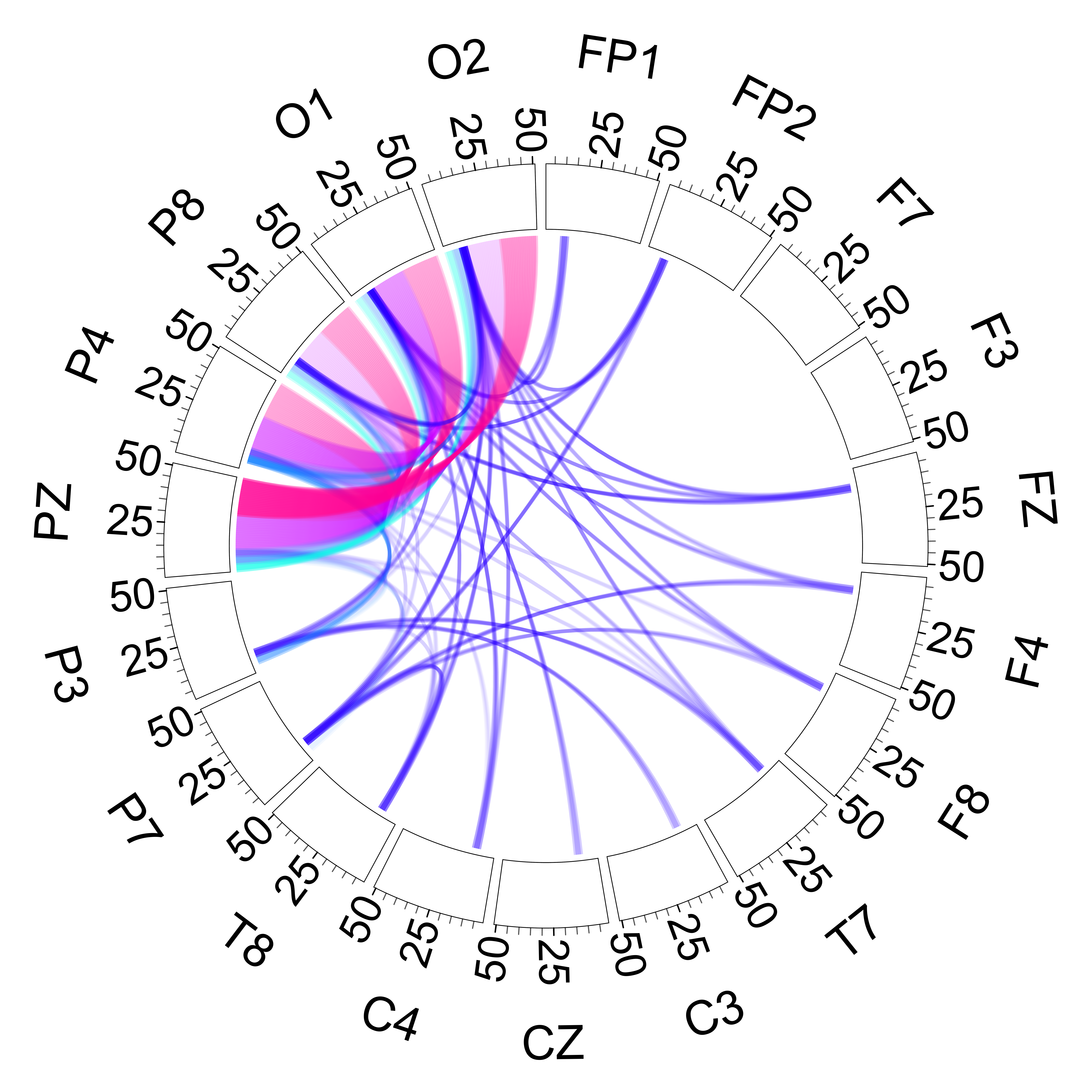}
\caption{SEED Dataset Electome \label{fig:electome_eeg}}
\end{subfigure}
\begin{subfigure}{0.32\textwidth}
\includegraphics[width=\textwidth]{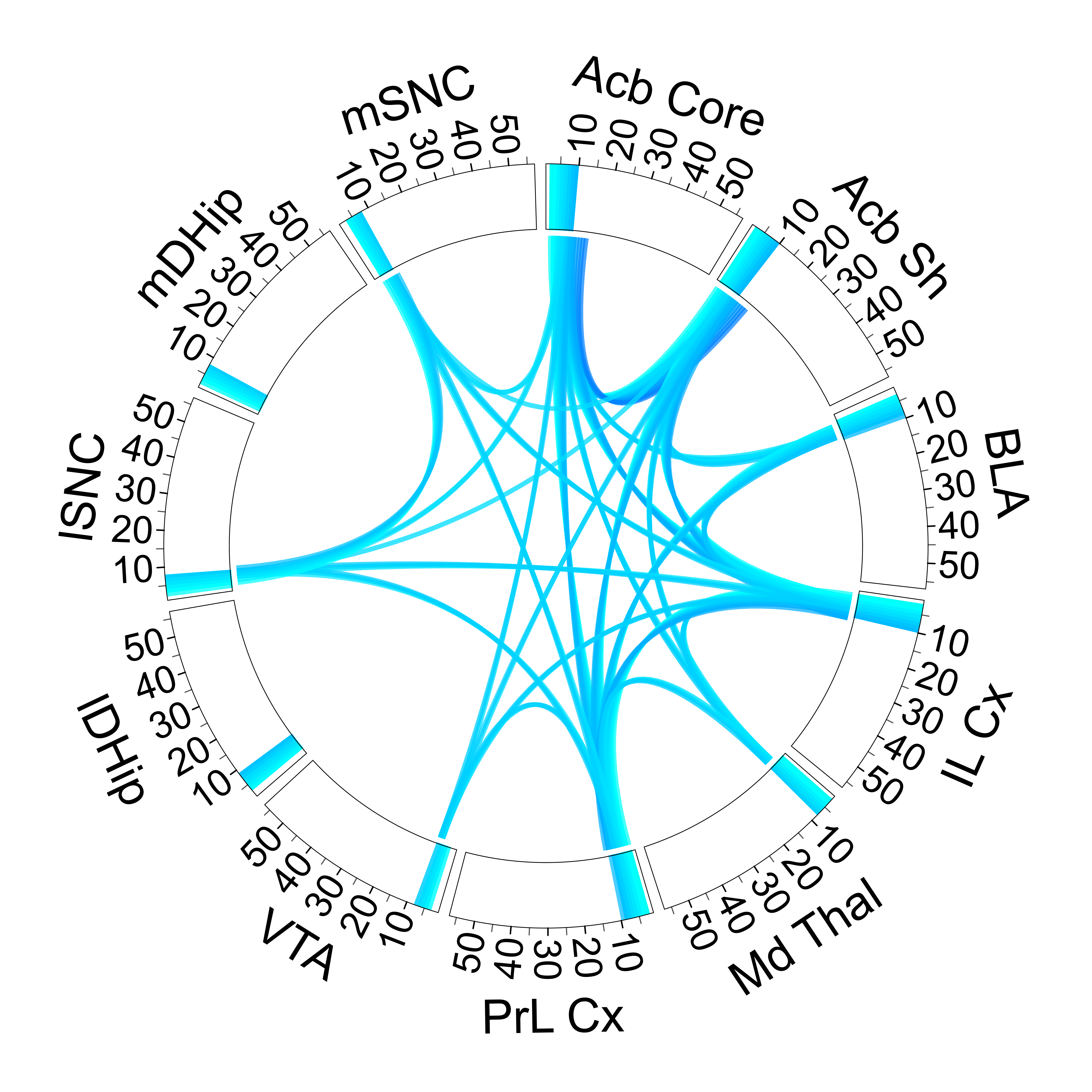}
\caption{Fully observed SOS-VAE \label{fig:electome_lfp_full} }
\end{subfigure}
\begin{subfigure}{0.32\textwidth}
\includegraphics[width=\textwidth]{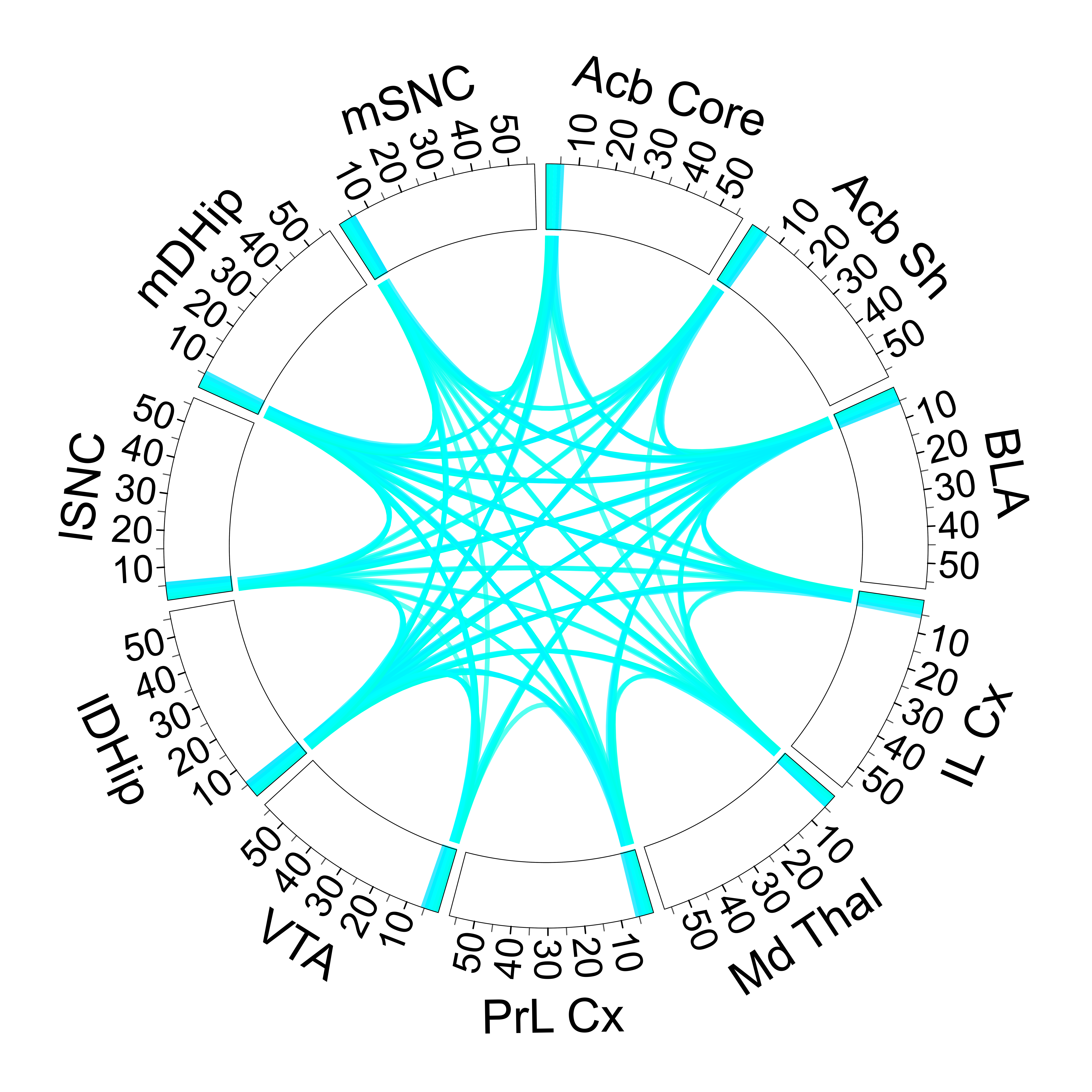}
\caption{Partially observed SOS-VAE \label{fig:electome_lfp_partial} }
\end{subfigure}
\caption{The single most predictive factor taken from the NMF decoder visualized as a network of brain regions. (a) SOS-DVAE model trained on SEED dataset. It uses 5 distinct frequency bands are represented here: Delta (1-4 \emph{Hz}; cyan), Theta (5-8 \emph{Hz}; blue), Alpha (9-12 \emph{Hz}; indigo), Beta (13-30 \emph{Hz}; violet), and Gamma (31-50 \emph{Hz}; magenta). (b) and (c) SOS-VAE model trained on LFP data with all and missing channels, respectively (see Appendix \ref{ssec:eMissingData}). The outermost set of labels are abbreviated names for each of the brain regions present in the recordings. The inner labels iterate over the modeled frequencies. Colored segments along the outer ``wheel'' of the image indicate that the factor represents signal power within that region and frequency range. Colored ``spokes'' between regions indicate that the factor represents coherence between those two regions at the associated frequency.\label{fig:electome_compare}}
\end{figure*}

A major goal of this work is to improve the utility of generative models for drawing scientific conclusions. We have shown that our proposed modifications to the SVAE produce generative models that more accurately explain variables of interest through a supervised task. For neural datasets such as this one, we can use the decoder parameters to draw scientific conclusions.
By using a non-negative matrix factorization model as our decoder, we can interpret the learned factors of the decoder as network factors of the electrical functional connectome (\emph{electome}) ~\citep{Gallagher2017Cross-spectralAnalysis, Talbot2020SupervisedActivity, Hultman2018Brain-wideVulnerability}.
To demonstrate the practical usefulness of this approach, we visualize the latent electome factor with the largest weight in the logistic regression classifier in Fig. \ref{fig:electome_eeg}.
This factor represents a network of brain regions defined by the power and coherence signatures we expect the network to produce.
This network is defined by nearly full-brain synchrony in the alpha band, whereas the other bands have localized coherence between PZ, P4, P8, O1, and O2.

\subsubsection{Decoding Behavioral Context from Local Field Potential Recordings}
\label{sec:lfps}
We next apply the proposed inference methods to a dataset of Local Field Potentials (LFPs) recorded from 11 different brain regions across 26 mice~\citep{Gallagher2017Cross-spectralAnalysis, Carlson2017Dynamically-timedPathway}.
Full brain region names are given in Supplemental Table \ref{st:lfp_channels}, with abbreviated names referenced in Fig. \ref{fig:electome_compare}.
Each mouse is recorded in three different behavioral contexts, which are thought to induce low, medium, and high levels of stress respectively.
As with the EEG dataset, recordings are divided into 1 second non-overlapping time windows.
Spectral power within each brain region and coherence between brain regions are calculated at frequencies from 1 \emph{Hz} to 56 \emph{Hz} in 1 \emph{Hz} increments for each time window \cite{Talbot2020SupervisedActivity}.
 We use a multinomial logistic regression for supervised classification of the behavioral context and evaluate the model performance via a
 5-fold cross-validation over mice.
 We report results for two different types of decoders, MLP and NMF, as in the previous sections.

For each inferred model we evaluate performance on the supervised task as well as the KL between posterior distributions if possible (see Table \ref{table:lfp-acc-auc}). 
As expected, SVAE and SOS-DVAE display comparable classification performance while SVAE-refit has near random performance, indicating that the performance of the SVAE model is not associated with the generative aspects of the model. 
Supplemental Figs. \ref{fig:kl_curves}e and \ref{fig:kl_curves}f visualize the tradeoff between predictive performance and KL for SDVAE and SOS-DVAE. The SOS-DVAE is more predictive at the same KL, demonstrating that it is guiding the generative model towards the classification goal. 

%Table \ref{table:lfp_missingdata} compares the prediction performance between SVAE and SOS-VAE, the results further confirmed that the SVAE fits a biased encoder which is not generalize well to unseen pattern.

Fig. \ref{fig:electome_lfp_full} shows the electome network with the largest weight in the classifier. The network is positively associated with the medium stress behavioral context, and demonstrates significant increases in connectivity between several brain regions in the low Gamma band of 30-50 \emph{Hz} and synchrony between an overlapping group of regions around 12\emph{Hz}.

%% file: texfiles/conclusions.tex
\section{Discussion and Conclusion}
\label{sec:discussion}

Generative latent variable models have great utility in scientific and clinical trial analysis to improve hypothesis generation and experimental design \citep{Rudin2019StopInstead,Rudin2019TheAnalysisc}. This scientific and clinical utility often depends on two goals, obtaining an accurate representation of the data and ensuring that the latent representation is predictive of an auxiliary task. A commonly used inference method, the SVAE, has been previously used to achieve both of these objectives. However, this objective function induces a previously-undetected bias in the encoder that hinders scientific utility.  We have shown, both on synthetic and real data, that the theoretical bias in SVAEs corresponds to a  detrimental impact on learned representations in practice.
To address this, we developed a novel inference technique that allows for supervision of an auxiliary task while maintaining a generative representation.
We demonstrate empirically that our novel inference technique achieves both scientific objectives without bias, and demonstrated the efficacy of the proposed methodology in relevant neuroscience applications. Furthermore, we have provided two relevant extensions to our methods that address critical needs in the neuroscience community.

We see two possible limitations of the proposed method. First, our models are designed for interpretable generative models commonly used in scientific analysis. This objective was not typically present in related work, which largely used the generative model as a regularization technique. When pursuing purely a predictive problem, a user can choose to use a much more complex yet difficult-to-interpret deep generative model \citep{kingma2014semi,maaloe2016auxiliary,li2017max,Finn2017Model-agnosticNetworks,Liu2019Self-SupervisedLearning}, making comparisons with the simpler, interpretable generative model learned via an  SOS-VAE irrelevant. Second, our contributions are applicable to models that include both a generative and a supervised component in the class of SAEs, whereas many supervised models have alternative inference strategies, such as \citep{Gallagher2017Cross-spectralAnalysis}.  

In conclusion, our developed inference techniques are highly relevant to scientific fields such as neuroscience that use latent variable models to design experiments and discover novel relationships in high-dimensional data. These techniques improve the scientific utility of these latent variable models by incorporating predictive information while maintaining a clear understanding of how manipulations of observed covariates will result in changes in the latent space. In the future, we will continue applying this to real-world scientific problems and work to build greater integration with standard black-box variational inference tools \citep{Ranganath2014BlackInference,Tran2017Edward:Criticism,Dillon2017TensorFlowDistributions}.

%These methodologies can increase the trustworthiness of the provided explanations of the model and facilitate an ``explanation-by-generation'' approach. In the future, we will continue applying this to real-world scientific problems and work to build greater integration with standard black-box variational inference tools \citep{Ranganath2014BlackInference,Tran2017Edward:Criticism,Dillon2017TensorFlowDistributions}.

%% file: texfiles/appendix.tex
\subsection{Derivation of the Fixed Points}
\label{ssec:aFixedPoints}

\subsubsection{Supervised Variational Autoencoder}

The objective of a SVAE for a single sample is 
\begin{equation}
    \calL_{\phi,\theta,\psi}= \E_{q_\phi(\s|\x)}\bigl[\log p_\theta(\x,\s) - \log q_\phi(\s|\x) + \lambda \log p_\psi(y|\s)\bigr].
\end{equation}
The gradients of $\theta$ and $\psi$ are straightforward. However, the gradients of $\phi$ are more difficult, as the ELBO expectation is taken with respect to a random sample from $q_\phi(\s|\x)$. However, if we express the random variable $\s\sim q_\phi(\s|\x)$ as a transformation of random variable $\epsilon$ given $\x$ and $\phi$, $g_\phi(\epsilon,\x)$, the distribution of $\epsilon$ will be independent of $\x$ and $\phi$. 
The gradients for $\theta$ are
\begin{equation}
\resizebox{\columnwidth}{!}{$
    \begin{split}
        \nabla_\theta \calL_{\phi,\theta,\psi}&=\nabla_\theta\E_{q_\phi(\s|\x)}\bigl[\log p_\theta(\x,\s) - \log q_\phi(\s|\x) +\\&\qquad\qquad\qquad\qquad \lambda \log p_\psi(y|\s)\bigr],\\
        &=\nabla_\theta\E_{p(\epsilon)}\bigl[\log p_\theta(\x,g_\phi(\epsilon,\x)) - \log q_\phi(g_\phi(\epsilon,\x)|\x) + \\&\qquad\qquad\qquad\qquad\lambda \log p_\psi(y|g_\phi(\epsilon,\x))\bigr],\\
        &=\E_{p(\epsilon)}\bigl[\nabla_\theta(\log p_\theta(\x,g_\phi(\epsilon,\x)) - \\&\qquad\;\;\log q_\phi(g_\phi(\epsilon,\x)|\x) + \lambda \log p_\psi(y|g_\phi(\epsilon,\x)))\bigr],\\
        &=\E_{p(\epsilon)}\bigl[\nabla_\theta\log p_\theta(\x,g_\phi(\epsilon,\x))\bigr],
    \end{split}
    $}
\end{equation}
while the gradients for $\psi$ are
\begin{equation}
\resizebox{\columnwidth}{!}{$
    \begin{split}
        \nabla_\psi \calL_{\phi,\theta,\psi}&=\nabla_\psi\E_{q_\phi(\s|\x)}\bigl[\log p_\theta(\x,\s) - \log q_\phi(\s|\x) +\\&\qquad\qquad\qquad\qquad \lambda \log p_\psi(y|\s)\bigr],\\
        &=\nabla_\psi\E_{p(\epsilon)}\bigl[\log p_\theta(\x,g_\phi(\epsilon,\x)) - \\&\qquad\qquad\log q_\phi(g_\phi(\epsilon,\x)|\x) + \lambda \log p_\psi(y|g_\phi(\epsilon,\x))\bigr],\\
        &=\E_{p(\epsilon)}\bigl[\nabla_\psi(\log p_\theta(\x,g_\phi(\epsilon,\x)) - \\&\qquad\qquad\log q_\phi(g_\phi(\epsilon,\x)|\x) + \lambda \log p_\psi(y|g_\phi(\epsilon,\x)))\bigr],\\
        &=\E_{p(\epsilon)}\bigl[\nabla_\psi\lambda \log p_\psi(y|g_\phi(\epsilon,\x)))\bigr],\\
        &=\lambda\E_{p(\epsilon)}\bigl[\nabla_\psi \log p_\psi(y|g_\phi(\epsilon,\x)))\bigr].\\
    \end{split}
    $}
\end{equation}
Finally, the gradients for $\phi$ are 
\begin{equation}
\resizebox{\columnwidth}{!}{$
    \begin{split}
        \nabla_\phi \calL_{\phi,\theta,\psi}&=\nabla_\phi\E_{q_\phi(\s|\x)}\bigl[\log p_\theta(\x,\s) - \log q_\phi(\s|\x) +\\&\qquad\qquad\qquad\qquad \lambda \log p_\psi(y|\s)\bigr],\\
        &=\nabla_\phi\E_{p(\epsilon)}\bigl[\log p_\theta(\x,g_\phi(\epsilon,\x)) - \log q_\phi(g_\phi(\epsilon,\x)|\x)\\&\qquad\qquad\qquad\qquad + \lambda \log p_\psi(y|g_\phi(\epsilon,\x))\bigr],\\
        &=\E_{p(\epsilon)}\bigl[\nabla_\phi(\log p_\theta(\x,g_\phi(\epsilon,\x)) - \\&\qquad\log q_\phi(g_\phi(\epsilon,\x)|\x) + \lambda \log p_\psi(y|g_\phi(\epsilon,\x)))\bigr].\\
    \end{split}
   $ }
\end{equation}
Thus, the fixed points for $\theta$ and $\psi$ are
\begin{equation}
    \begin{split}
0&=\E_{p(\epsilon)}\bigl[\nabla_\theta\log p_\theta(\x,g_\phi(\epsilon,\x))\bigr],\\
0&=\E_{p(\epsilon)}\bigl[\nabla_\psi \log p_\psi(y|g_\phi(\epsilon,\x)))\bigr],
    \end{split}
\end{equation}

and the fixed point for $\phi$ is 
\begin{equation}
\begin{split}
  -\lambda \E_{p(\epsilon)}\bigl[&\nabla_\phi\log p_\psi(y|g_\phi(\epsilon,\x)))\bigr] = \E_{p(\epsilon)}\bigl[\nabla_\phi(\log p_\theta(\x,\\&g_\phi(\epsilon,\x)) - \log q_\phi(g_\phi(\epsilon,\x)|\x)\bigr].   
\end{split}
\end{equation}

\subsubsection{Second-Order Supervision}
The second-order supervision objective recouples the encoder and decoder by forcing the encoder to approximate solely the generative posterior. Mathematically this objective is
\begin{equation}
\begin{split}
    \textstyle \max_{\psi,\theta}\;\; &\E_{q_\phi(\s|\x)}\bigl[\log p_\theta(\x,\s)  + \lambda \log p_\psi(y|\s) \bigr]\\
    \textrm{s.t.}\;\; \phi &= \textstyle \argmax_{\phi'} \E_{q_{\phi'}(\s|\x)}\bigl[\log p_\theta(\x,\s)-\log q_{\phi'}(\s|\x)\bigr].
\end{split}
\end{equation}

The gradients for $\psi$ follow almost identically to the SVAE as
\begin{equation}
\resizebox{\columnwidth}{!}{$
    \begin{split}
        \nabla_\psi \calL_{\theta,\psi}&=\nabla_\psi\E_{q_\phi(\s|\x)}\bigl[\log p_\theta(\x,\s)  + \lambda \log p_\psi(y|\s)   \bigr],\\
        &=\nabla_\psi\E_{p(\epsilon)}\bigl[\log p_\theta(\x,g_\phi(\epsilon,\x)) + \lambda \log p_\psi(y|g_\phi(\epsilon,\x)) \bigr],\\
        &=\E_{p(\epsilon)}\bigl[\nabla_\psi(\log p_\theta(\x,g_\phi(\epsilon,\x)) + \lambda \log p_\psi(y|g_\phi(\epsilon,\x)))\bigr],\\
        &=\E_{p(\epsilon)}\bigl[\nabla_\psi\lambda \log p_\psi(y|g_\phi(\epsilon,\x)))\bigr],\\
        &=\lambda\E_{p(\epsilon)}\bigl[\nabla_\psi \log p_\psi(y|g_\phi(\epsilon,\x)))\bigr].\\
    \end{split}
    $}
\end{equation}
The gradients taken with respect to $\phi$ are done wrt the constraint as 
\begin{equation}
    \begin{split}
        \nabla_\phi &=\nabla_\phi\E_{q_{\phi}(\s|\x)}\bigl[\log p_\theta(\x,\s)-\log q_{\phi}(\s|\x)\bigr],\\
        &=\nabla_\phi\E_{p(\epsilon)}\bigl[\log p_\theta(\x,g_\phi(\epsilon,\x))-\log q_{\phi}(g_\phi(\epsilon,\x)|\x)\bigr],\\
        &=\E_{p(\epsilon)}\bigl[\nabla_\phi(\log p_\theta(\x,g_\phi(\epsilon,\x))-\log q_{\phi}(g_\phi(\epsilon,\x)|\x))\bigr].\\
    \end{split}
\end{equation}

Finally, the gradients with respect to $\theta$ are 
\begin{equation}
\resizebox{\columnwidth}{!}{$
    \begin{split}
        \nabla_\theta \calL_{\theta,\psi}&=\nabla_\theta\E_{q_\phi(\s|\x)}\bigl[\log p_\theta(\x,\s)  + \lambda \log p_\psi(y|\s)   \bigr],\\
        &=\nabla_\theta\E_{p(\epsilon)}\bigl[\log p_\theta(\x,g_\phi(\epsilon,\x)) + \lambda \log p_\psi(y|g_\phi(\epsilon,\x)) \bigr],\\
        &=\E_{p(\epsilon)}\bigl[\nabla_\theta(\log p_\theta(\x,g_\phi(\epsilon,\x)) + \lambda \log p_\psi(y|g_\phi(\epsilon,\x)))\bigr].
    \end{split}
    $}
\end{equation}
This implies that the fixed points of the SOS-VAE for $\phi$ and $\psi$ are 
\begin{equation}
    \begin{split}
0&=\E_{p(\epsilon)}\bigl[\nabla_\phi(\log p_\theta(\x,g_\phi(\epsilon,\x))-q(g_\phi(\epsilon,\x)|\x))\bigr],\\
0&=\E_{p(\epsilon)}\bigl[\nabla_\psi \log p_\psi(y|g_\phi(\epsilon,\x)))\bigr],
    \end{split}
\end{equation}
while the fixed point for $\theta$ is 
\begin{equation}
    \begin{split}{
 -\lambda &\E_{p(\epsilon)}\bigl[\nabla_\theta\log p_\psi(y|g_\phi(\epsilon,\x))\bigr] =\\ &\E_{p(\epsilon)}\bigl[\nabla_\theta(\log p_\theta(\x,g_\phi(\epsilon,\x))\bigr].
    }\end{split}
\end{equation}

\subsubsection{Gradient Update Derivations}
\label{ssec:aGradient}

As mentioned previously, $\nabla_\theta\log p_\psi(y|g_\phi(\epsilon,\x))$ is strange given that $\theta$ and $\phi$ are both variables in common implementations. However, the constraint induces dependence on $\phi$. To evaluate the gradient on $\theta$, note that the total derivative estimate for $\theta$ is 
\begin{equation}
    \frac{\partial\mathcal{L}}{\partial\theta}= \frac{\partial \log p_\theta(\x,\s)}{\partial\theta} + \frac{\partial \log p_\theta(\x,\s)}{\partial\phi}\frac{\partial\phi}{\partial\theta} + \frac{\partial \log p_\psi(y|\s)}{\partial\phi}\frac{\partial\phi}{\partial\theta}.
\end{equation}
The first term is trivial to implement in modern software packages. When the constraint on $\phi$ is approximately satisfied $\partial\phi/\partial\theta$ is small and can be ignored.  While this term is not close to zero when the networks are initialized, it is still small comparitively and can be reasonably ignored. The third term can be approximately evaluated using the second-order trick described in \citep{Finn2017Model-agnosticNetworks}.

\SetKwInput{kwInit}{Initialize}
\begin{algorithm}[t]
\SetAlgoLined
\KwIn{$\{\x_1,...,\x_N\}\in\mathbb{R}^p$, $\{y_1,...,y_N\}\in\mathcal{Y}$.}
\kwInit{Network parameters: $\phi_1$, $\phi_1^+$, $\phi_2,\theta,\psi$; learning rate: $\alpha, \beta$; weights: $\lambda,\mu$.}
\For{epoch in iterations}{
    $(\x_i,y_i), i\in\{1,...,N\}$ \algorithmiccomment{Training data batch}\\
    $\eta\sim$$p_\theta(s)$,
    %$\eta\sim$$N$$(0,I)$, 
    $\epsilon_i\sim$$N(0,I)$ \algorithmiccomment{Latent space definition}\\
    $\theta^+\gets\theta+\alpha\nabla_\theta(\log p_\theta(\x_i,g_{\phi_1}(\epsilon_i,\x_i))- KL(\epsilon_i,\eta))$\algorithmiccomment{Step 1: decoder update wrt VAE}\\
    $\phi_1^+\gets\phi_1+\alpha\nabla_{\phi_1}(\log p_\theta(\x_i,g_{\phi_1}(\epsilon_i,\x_i))- KL(\epsilon_i,\eta))$\algorithmiccomment{Step 2: Update  $\phi_1$}\\
    $s_1\sim{q_{\phi_1^+}(\s|\x_i)}\;\;\;s_2\sim{q_{\phi_2}(\s|\x_i)}$\algorithmiccomment{Step 3: Sample latent spaces}\\
    $\psi^+\gets\psi+\alpha\nabla_\psi(\lambda\log p_\psi(y_i|g_{\phi_2}(\epsilon_i,\x_i)) - \mu KL(s_2,s_1)$\algorithmiccomment{Step 4: Standard update to classifier}\\
    $\phi_2^+\gets\phi_2+\alpha\nabla_{\phi_2}(\lambda\log p_\psi(y_i|g_{\phi_2}(\epsilon_i,\x_i)))+\mu KL(s_2,s_1))$\algorithmiccomment{Standard update to encoder $q_{\phi_2}(\s|\x)$.}\\
    
    $\theta^{++}\gets\theta^+-\beta\nabla_{\theta^+}(\lambda\log p_\psi(y_i|g_{\phi_1^+}(\epsilon_i,\x_i))$\algorithmiccomment{Step 5: Second-order update to the decoder}\\
    $\psi\leftarrow \psi^+$, $\phi_1\leftarrow \phi_1^+$, $\phi_2\leftarrow \phi_2^+$, $\theta\leftarrow \theta^{++}$\algorithmiccomment{Update model parameters}
    }
\caption{\label{alg:sosdvae}Second-Order Supervision Double VAE (SOS-DVAE)}
\end{algorithm}

\begin{algorithm}[ht]
\SetAlgoLined
\KwIn{$\{\x_1,...,\x_N\}\in\mathbb{R}^p$, $\{y_1,...,y_N\}\in\mathcal{Y}$.}
\kwInit{Network parameters: $\phi_1$, $\phi_2,\theta,\psi$; learning rate: $\alpha$; weights: $\lambda,\mu,\eta$.}
\For{epoch in iterations}{
    $(\x_i,y_i), i\in\{1,...,N\}$ \algorithmiccomment{Training data batch}\\
    $\eta\sim$$p_\theta(s)$,
    %$\eta\sim$$N$$(0,I)$, 
    $\epsilon_i\sim$$N(0,I)$ \algorithmiccomment{Latent space definition}\\
    $\theta^+\gets\theta+\alpha\nabla_\theta(\log p_\theta(\x_i,g_{\phi_1}(\epsilon_i,\x_i))- KL(\epsilon_i,\eta))$\algorithmiccomment{Step 1: Update decoder wrt VAE}\\
    $\phi_1^+\gets\phi_1+\alpha\nabla_{\phi_1}(\log p_\theta(\x_i,g_{\phi_1}(\epsilon_i,\x_i))- KL(\epsilon_i,\eta))$\algorithmiccomment{Step 2: Update $\phi_1$}\\
    $s_1\sim{q_{\phi_1^+}(\s|\x_i)}\;\;\;s_2\sim{q_{\phi_2}(\s|\x_i)}$\algorithmiccomment{Step 3: Sample latent spaces}\\
    $\psi^+\gets\psi+\alpha\nabla_\psi(\lambda\log p_\psi(y_i|g_{\phi_2}(\epsilon_i,\x_i)) - \mu KL(s_2,s_1)$\algorithmiccomment{Step 4: Update classifier}\\
    $\phi_2^+\gets\phi_2+\alpha\nabla_{\phi_2}(\lambda\log p_\psi(y_i|g_{\phi_2}(\epsilon_i,\x_i)))+\mu KL(s_2,s_1))$ \algorithmiccomment{Step 5: Update $\phi_2$}\\
    $\psi\leftarrow \psi^+$, $\phi_1\leftarrow \phi_1^+$, $\phi_2\leftarrow \phi_2^+$, $\theta\leftarrow \theta^{+}$\algorithmiccomment{Update model parameters}
    }
\caption{\label{alg:sdvae} Supervised Double VAE (SDVAE)}
\end{algorithm}

%\begin{algorithm}[h]
%\SetAlgoLined
%\KwIn{$\{x_1,...,x_N\}\in\mathbb{R}^p$, $\{y_1,...,y_N\}\in\mathcal{Y}$.}
%\kwInit{Network parameters: $\phi_1$, $\phi_2,\theta,\psi$; learning rate: $\alpha$; weights: $\lambda,\mu,\eta$.}
%\For{epoch in iterations}{
%    \# Fetch one batch of training data\;\\
%    $(\x_i,y_i), i\in\{1,...,N\}$\;\\
%    \# Define latent spaces\;\\
%    $q_0\gets$$N$$(0,I)$, $q_1\gets$$f_\phi$$(\x_i)$\;\\
%    \# Standard update on the decoder w.r.t. VAE loss\;\\
%    $\theta^+\gets\theta-\alpha\nabla_\theta(\lossx(\x_i,g_\theta(f_{\phi_1}(\x_i))+\eta KL(q_1,q_0))$\;\\
%    \# Standard update on encoder $f_{\phi_1}$\;\\
%    $\phi_1^+\gets\phi_1-\alpha\nabla_{\phi_1}(\lossx(\x_i,g_\theta(f_{\phi_1}(\x_i))+\eta KL(q_1,q_0))$\;\\
%    \# Calculate the latent space\;\\
%    $q_1\gets{f_{\phi_1^+}(\x_i)}$\;\\
%    $q_2\gets{f_{\phi_2}(\x_i)}$\;\\
%    \# Standard update on classifier w.r.t. classification loss and the discrepancy between the two encoders\;\\
%    $\psi^+\gets\psi-\alpha\nabla_\psi(\lambda\lossy(y_i,h_\psi(f_{\phi_2}(\x_i)))+\mu KL(q_2,q_1))$\;\\
%    \# Standard update to encoder $f_{\phi_2}$\;\\
%    $\phi_2^+\gets\phi_2-\alpha\nabla_{\phi_2}(\lambda\lossy(y_i,h_\psi(f_{\phi_2}(\x_i)))+\mu KL(q_2,q_1))$\;\\
%    \# Update model parameters\;\\
%    $\psi\leftarrow \psi^+$, $\phi_1\leftarrow \phi_1^+$, $\phi_2\leftarrow \phi_2^+$, $\theta\leftarrow \theta^{+}$\;
%    }
%\caption{\label{alg:sdvae} Supervised Double VAE (SDVAE)}
%\end{algorithm}

\subsection{Details on Algorithmic Settings and Learning}
\label{ssec:aSettings}

Algorithm \ref{alg:sosdvae} provides detailed pseudocode for the  Second-Order Supervision Double VAE (SOS-DVAE) matching the 5 steps in Figure \ref{fig:overview} (in the main manuscript). If we ignore step 5 (the second-order update on the decoder), the algorithm is the Supervised Double VAE (SDVAE) as in Algorithm \ref{alg:sdvae}. For simplicity, the standard update to encoder $q_{\phi_2}(\s|\x)$ (line \#8 in Algorithm \ref{alg:sosdvae}) is omitted in Figure \ref{fig:overview} in the main manuscript.

For the reported experiments, all models use a Gaussian distribution as the prior on the latent space ($q_0\gets$$N$$(0,1)$). The Adam optimizer is used for the gradient optimization. Parameters used for each dataset are listed in Table \ref{st:params}. The learning rate $\beta$ used in SOS-VAE (Algorithm \ref{alg:sosvae} in the main manuscript) and SOS-DVAE (Algorithm \ref{alg:sosdvae}) is initialized with the same value as $\alpha$ (Table \ref{st:params}), and decays by half after 50 epochs (Table \ref{st:params}). 

Code and instructions have been included with the submission to recreate the experimental results on the MNIST and SEED datasets.

\subsection{Incorporating Missing Data}
\label{ssec:eMissingData}

It is often scientifically useful to combine datasets that have overlapping but distinct measurements. In neural recording applications, this frequently occurs when multiple experiments that record from distinct but largely overlapping brain regions are combined to increase sample size. A common approach for combining these distinct datasets is to treat synthesis as a missing data problem. Generative models with Bayesian inference provide a natural method to impute the missing covariates and extract scientific conclusions from the decoder parameters \citep{soudry2015efficient,gelman1995bayesian}. In our motivating application, we will combine datasets that recorded LFPs from distinct brain regions with some overlap and use a VAE framework for efficient and predictive inference.

We assume that the data come from $T$ experiments, where $\{\x_1^t,...,\x_{N_t}^t\}\in\R^{p_t}$ are $N_t$ independent samples from the $t$-th experiment. We let $\x\in\R^{q}$ represent the entire (but not completely observed) data from all recorded regions, where $\max(p_1,\dots,p_T)\le q<p_1+\dots+p_T$. %We defer all proofs and derivations to the Supplemental Section \ref{ssec:aFixedPoints} for brevity.

We learn $T$ encoder networks $\{q_{\phi^t}(\s|\x^t)\}$ to approximate the $t$-th posterior $p_\theta(\s|\x^t)$ conditioned on the observed covariates for each experiment.  At test time on a new experiment with different regions observed, we would need to approximate the posterior, which we can do by learning one additional encoder. Unfortunately, the inherent issues of deviation in encoders arises if we maximize an adapted version of \eqref{eq:svae}.  We show empirically that this approach is flawed in our applications since the posterior distribution is inconsistent.  Fortunately, the objective of \eqref{eq:sosvae} can easily be adapted to handle the missing data as
\begin{equation}\label{eq:sosvae_m}
\begin{split}
\textstyle 
    \max_{\psi,\theta}\;\; &
    \textstyle \sum_{t=1}^T\E_{q_{\phi^t}(\s|\x^t)}\bigl[\log p_\theta(\x^t,\s)  + \lambda \log p_\psi(y|\s) \bigr]\\
    \textrm{s.t.}\;\; &\textstyle \phi^t = \argmax_{\phi'} \E_{q_{\phi'}(\s|\x^t)}\bigl[\log p_\theta(\x^t,\s)-\\
    &\quad\quad \quad\quad \log q_{\phi'}(\s|\x^t)\bigr]\text{ for }t\in1,\dots,T.
\end{split}
\end{equation}

The constraints in \eqref{eq:sosvae_m} enforce that each encoder performs proper posterior inference, ensuring that the latent space obtained via each encoder reconstructs the the observed data while predicting the unobserved covariates. Dependence between the encoders is obtained through a shared $\theta$ between all experiments. Given a test sample $\x^*$ with a new set of observed regions, the posterior can be approximated by learning a new encoder with parameters $\phi^*$ using the objective
\begin{equation}
\textstyle
\phi^* = \argmax_{\phi'} \E_{q_{\phi'}(\s|\x^*)}\bigl[\log p_\theta(\x^*,\s)-\log q_{\phi'}(\s|\x^*)\bigr].
\end{equation}
If the training encoders approximate exclusively a generative model (a VAE), this objective will be consistent. However, in an SVAE formulation this approach will approximate the ``refit'' encoder defined by the generative model. If the predictive capability of the SVAE is largely obtained via the deviation, this approximation will lose most of the predictive ability. Our methodology will be resilient to such problems, as the supervision is induced in the decoder parameters $\theta$, which are used to train the new encoder.

%If the training encoders approximate the generative model this will be consistent; however, in an SVAE the training encoders are biased and this approximation will lose most of the predictive ability.

\subsection{Implementation Details}
\label{sup:implementation}
\label{sup:implementationalt}
For all models, the encoder uses a single hidden layer MLP with 512 nodes. Two decoders are investigated: ($i$) a single layer MLP with 512 nodes, and ($ii$) a Non-negative Matrix Factorization (NMF) decoder.  In the NMF decoder, the latent space is mapped to non-negative values through a softplus non-linearity, and then a non-negative linear mapping is learned to project to the outputs.

All models are implemented using Pytorch. The experiments are run on a cluster with a Red Hat Enterprise Linux 7 operating system and a range of Nvidia GPUs, including TitanXPs and RTX2080Tis. Training parameters used for the experiments for each dataset are summarized in Table \ref{st:params}.

\begin{table}[h]\centering
    \caption{Training parameters for each dataset. $\lambda$ and $\eta$ are weighting parameters, and $\alpha$ is the learning rate.}
    \resizebox{\columnwidth}{!}{
    \begin{tabular}{cccccccc}
    \hline\hline
    Dataset & Feature size&Categories&Batch size&$\alpha$&$\lambda$ &$\eta$ \\
    \hline
    MNIST &28x28 &10 &128 &1e-3 &1e-3 &0.1\\ 
    LFP &1x3696 &3 &100 &1e-4 &100 &10 \\
    SEED &1x950 &3 &64 &1e-5 &1 &1e-4 \\
    \hline
    \end{tabular}
    }
    \label{st:params}
\end{table}

All hyperparameters were selected in the same way. For a standard SVAE and SOS-VAE, the $\lambda$, learning rate, and batch size are important hyperparameters. In the SDVAE and SOS-DVAE models an extra hyperparameter is a scaling factor $\mu$ balancing the Kullback-Liebler (KL) divergence with the other terms of the objective function. These parameters are tuned in 3 steps: 1) a sparse search with 13 values in a bigger range to locate smaller ranges that provide better prediction performance with 10-fold cross-validation, 2) shrink the searching range and add new values in the new searching range that obtains better performance, 3) run multiple times and chose the parameters that offers the best performance. For $\lambda$, the initial search range is [1e-8, 1e8]. For $\mu$, the initial search ranging [1e-5,5] to investigate the trade off between generation and inference for each dataset using MLP decoder (see Figure \ref{fig:kl_curves}). Note that we used a single hidden layer MLP to demonstrate the broad applicability of the methods, and that the modification to a more complex architecture is straightforward to implement.

\subsubsection{MNIST}
The MNIST contains 60,000 training images and 10,000 test images, which we concatenate together into one dataset for a 10-fold cross-validation with random splits. No data augmentation is applied to the dataset, since our goal is to compare the performance among different models, not to pursue the best prediction.

\subsubsection{Local Field Potential (LFP)}
\label{sup:lfp}
The Local Filed Potentials (LFPs) are recorded from 11 different brain regions (see Table \ref{st:lfp_channels}) for each mouse \cite{Carlson2017Dynamically-timedPathway}. Recordings are split into 1\emph{s} intervals, each with an associated genotype and condition label. We estimate the spectral power features for each time interval using Welch's method \citep{Welch1967ThePeriodograms} and mean squared coherence between pairs of brain regions \citep{Rabiner1975} as measures of frequency-resolved synchrony within and between regions, respectively. These features are calculated at 1 \emph{Hz} intervals from 1 \emph{Hz} to 56 \emph{Hz}, yielding a 3696 dimensional observation space. 

The model for the outcome given the latent factors naturally lends itself to multinomial regression for the prediction of low-, medium-, and high-stress contexts corresponding to experimental conditions of home cage, open field, and tail suspension respectively. The NMF decoder naturally lends itself as a biologically interpretable model for neural electrophysiology \citep{Talbot2020SupervisedActivity}. It views each observation as a positive sum of non-negative features, which matches the biological assumption that no network of neural activity can be negatively activated or be associated with with negative power or coherence. We also use an MLP decoder to demonstrate the broad applicability of these methods.

\begin{table}[h]\centering
    \caption{The 11 brain regions in the LFP dataset.}
    \begin{tabular}{cll}
    \hline\hline
    ID & Abbreviation & Full name \\
    \hline
    1 & Acb Core & Nucleus Accumbens Core \\ 
    2 & Acb Sh & Nucleus Accumbens Shell \\
    3 & BLA & Basolateral Amygdala \\
    4 & IL Cx & Infralimbic Cortex \\
    5 & Md Thal & Mediodorsal Nucleus of the Thalamus \\
    6 & PrL Cx & Prelimbic Cortex \\
    7 & VTA & Ventral Tegmental Area \\
    8 & lDHip & Lateral Dorsal Hippocampus \\
    9 & lSNC & Lateral Substantia Nigra Pars Compacta \\
    10 & mDHip & Medial Dorsal Hippocampus \\
    11 & mSNC & Medial Substantia Nigra Pars Compacta \\
    \hline
    \end{tabular}
    \label{st:lfp_channels}
\end{table}

\label{ssec:aSOS_missingData}

\subsubsection{Electroencephalography(EEG)}
\label{sup:eeg}
\begin{figure}[t]
    \centering
    \includegraphics[width=0.7\columnwidth]{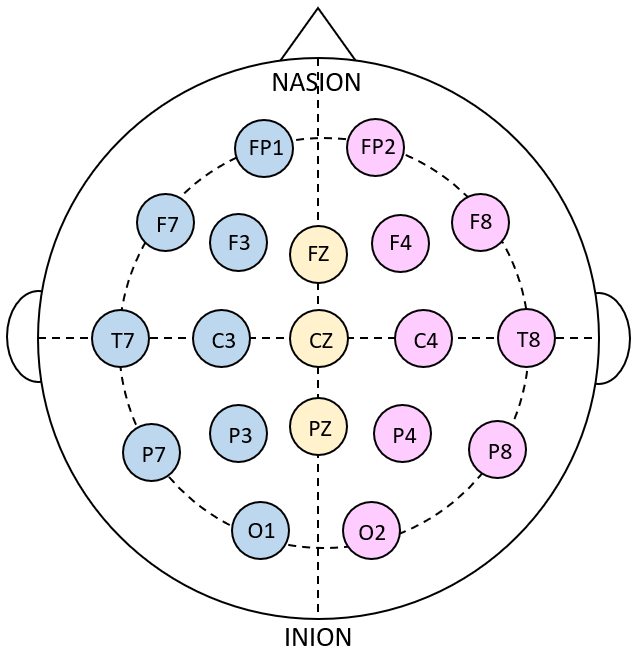}
    \caption{Locations of a subset of 19 electrodes for the SEED dataset.}
    \label{fig:EEG_electrodes}
\end{figure}

For the electroencephalography (EEG) recordings of the SEED dataset \citep{duan2013differential,zheng2015investigating}, we use a subset of 19 electrodes (as shown in Figure \ref{fig:EEG_electrodes}) that approximate the whole head to simplify visualizations, but it would be a straightforward extension to use the full set of 62 electrodes. Power and coherence features are calculated for the 19 electrodes in 1 second time windows at the following frequency bands: 1-4, 5-8, 9-12, 13-30, and 31-50 \emph{Hz}, corresponding to the Delta, Theta, Alpha, Beta, and Low Gamma, respectively. The resulting power and coherence features are flattened and concatenated to yield a 950 dimensional observation space. We use the MNE package (an open-source tool available at: \href{https://mne.tools/stable/index.html}{https://mne.tools/stable/index.html}) to extract the power and coherent features for this EEG dataset.

\begin{figure*}[t]
    \centering
    \includegraphics[width=\textwidth]{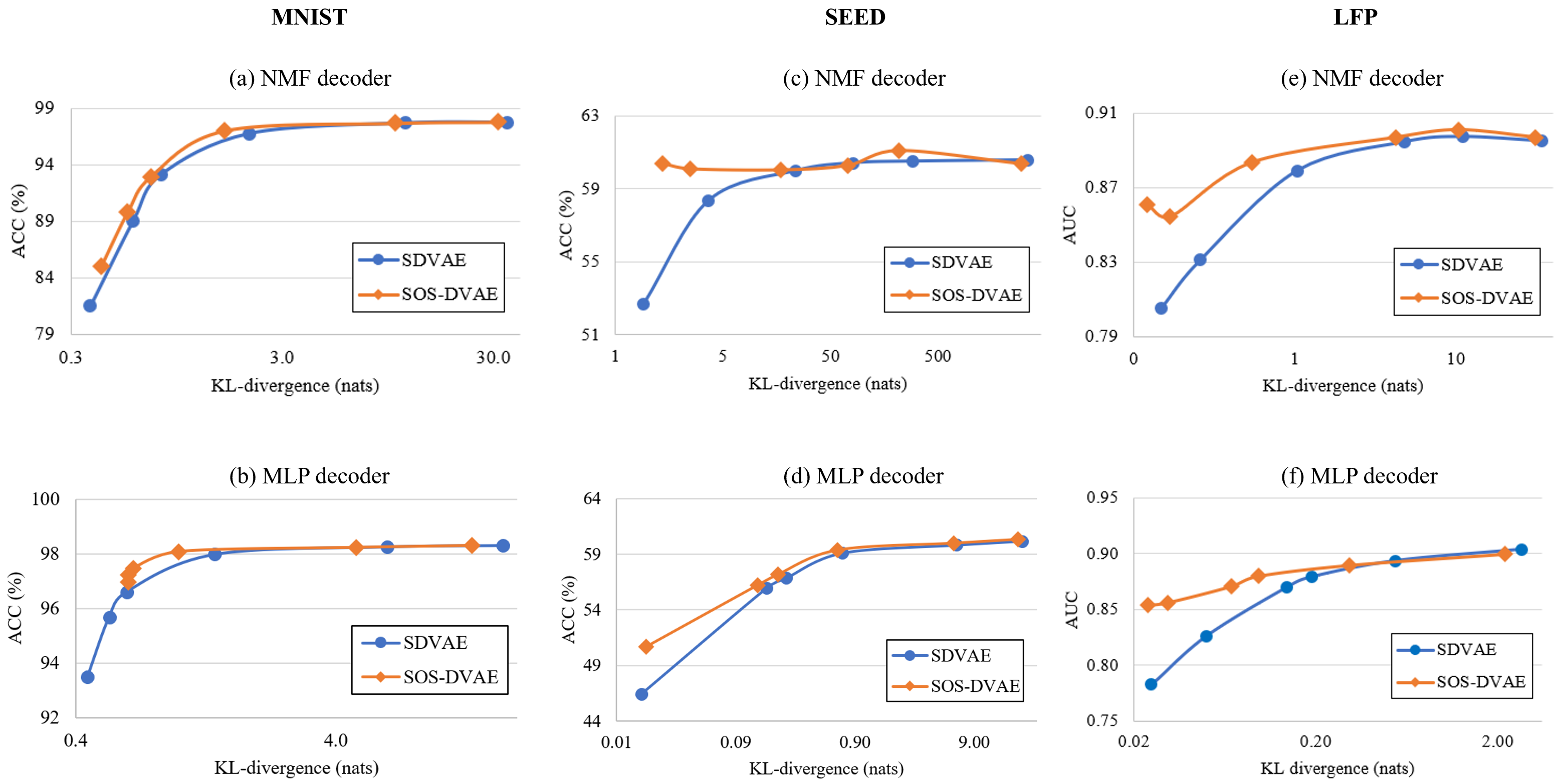}
    \caption{Prediction performance on MNIST (left), SEED (middle), and LFP (right) datasets using NMF decoder (top row) and MLP decoder (bottom row).}
    \label{fig:kl_curves}
\end{figure*}

\begin{figure*}[t]
    \centering
    \includegraphics[width=\textwidth]{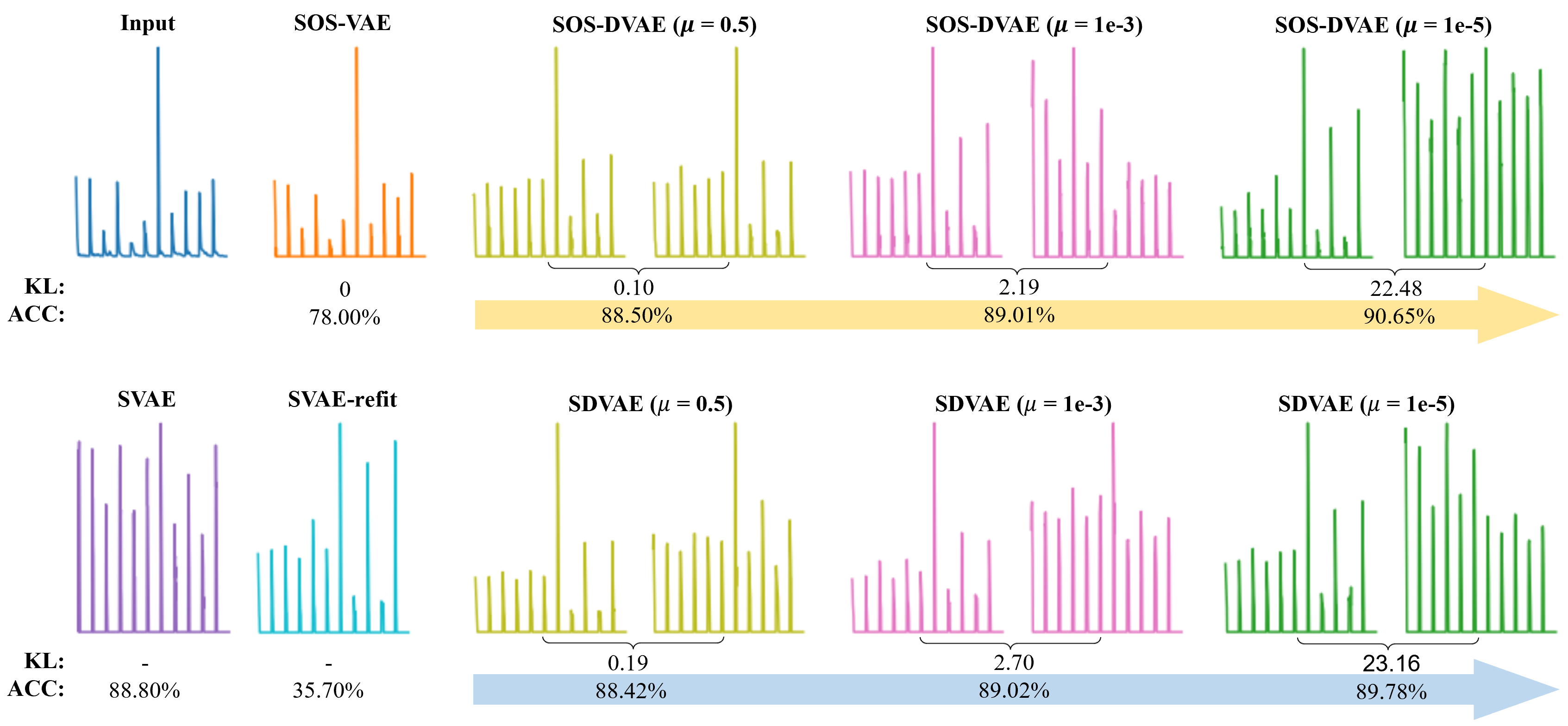}
    \caption{Visualization of reconstructed power features on LFP dataset using MLP decoder. In the bottom left, we show the results for an SVAE model before and after (SVAE-refit) updating the encoder to prioritize the generative model. The results of SOS-DVAE (above yellow arrow) and SDVAE (above blue arrow) models are shown with 3 different KL-divergences (controlled by $\mu$ as detailed in Section \ref{ssec:eRelaxing} and in Algorithm \ref{alg:sosdvae}). For each $\mu$, the left and right pictures are reconstructed from the inference encoder $q_{\phi_1}(\cdot)$ and the classification $q_{\phi_2}(\cdot)$ encoder, respectively. Bigger $\mu$ (smaller KL-divergence) yields similar reconstruction from the two encoders, vice versa. The numbers under the picture are KL-divergence in nats and prediction AUC. The models used in this figure is trained on a single split of training set, and the reconstruction is for a random input in the hold-out test set.}
    \label{fig:LFP_generative}
\end{figure*}

\begin{figure*}[t]
    \centering
    \includegraphics[width=\textwidth]{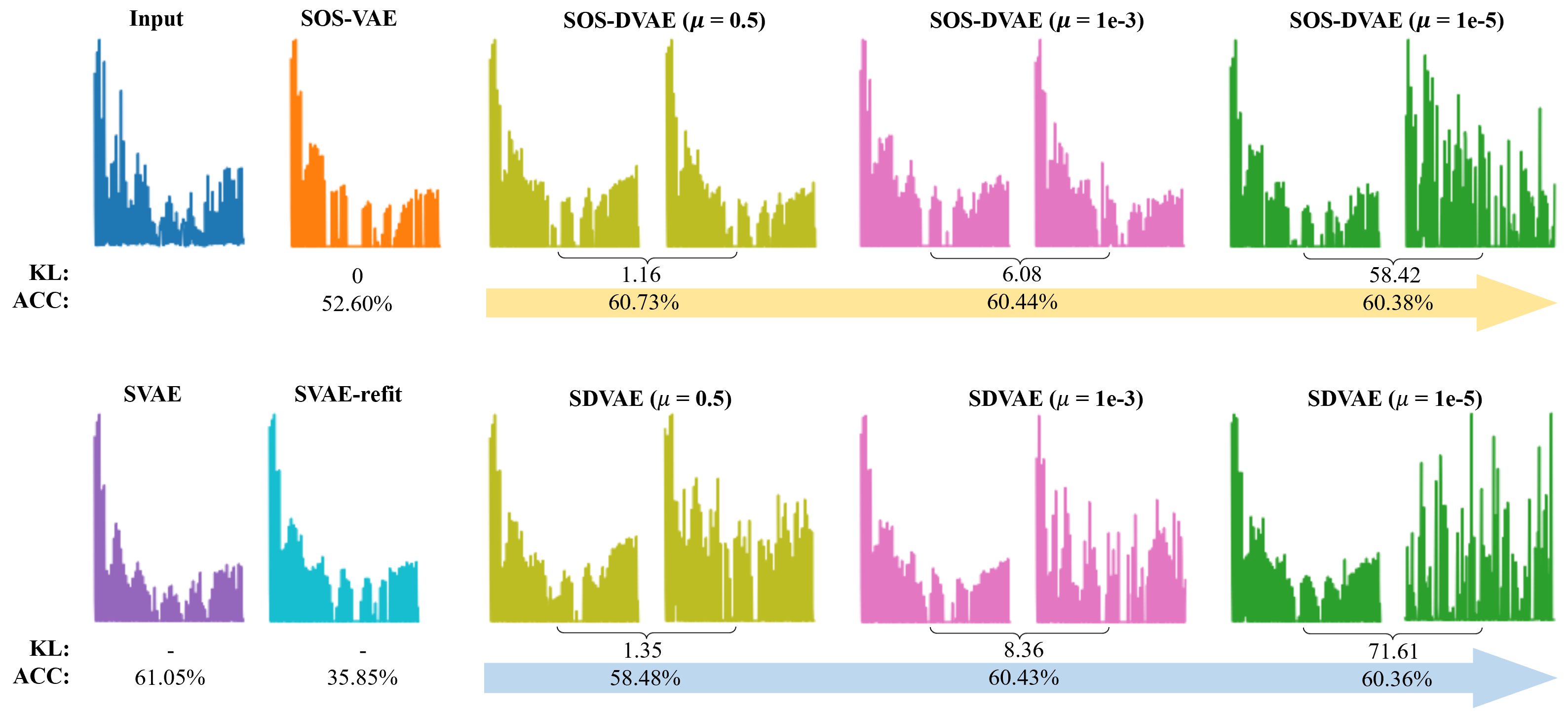}
    \caption{Visualization of reconstructed power features on SEED dataset using an MLP decoder. In the bottom left, we show the results for an SVAE model before and after (SVAE-refit) updating the encoder to prioritize the generative model. The results of SOS-DVAE (grouped by the yellow arrow) and SDVAE (grouped by the blue arrow) models are shown with 3 different KL-divergences (controlled by $\mu$ in Algorithm \ref{alg:sosdvae}). For each $\mu$, the left and right pictures are reconstructed from the inference encoder $q_{\phi_1}(\cdot)$ and the classification $q_{\phi_2}(\cdot)$ encoder, respectively. Bigger $\mu$ (smaller KL-divergence) yields similar reconstruction from the two encoders, vice versa. The numbers under the picture are KL-divergence in nats and prediction accuracy in percentage. The models used in this figure is trained on a single split of training set, and the reconstruction is for a random input in the hold-out test set.}
    \label{fig:seed_generative}
\end{figure*}

\subsection{SOS-DVAE Improves Robustness to Missing Data}
%An important application of the proposed method is to learn a consistent decoder from incomplete dataset.
One of the appealing novelties of our method is that it addresses missing data issues that often arise when combining recordings from multiple scientific experiments. To demonstrate the concept of such application, we first generate a synthetic dataset by randomly removing 3 brain regions in each mouse in the LFP dataset (described in Section \ref{sec:lfps}), and then we apply the extension of our proposed method to handling such missing data. Finally, we compare the trained models with that trained on full channels as shown in Figures \ref{fig:electome_lfp_full} and  \ref{fig:electome_lfp_partial}. 

In practice, data may not be collected consistently, resulting in missing portions of the data. For example, LFP recordings such as the ones used above often suffer from overly noisy channels that prevent signal from being observed in one or more of the brain regions in the study. We simulate this scenario by randomly removing 3 brain regions in each mouse in the LFP dataset from Section \ref{sec:lfps}, and we apply the missing data methodology from Section \ref{ssec:eMissingData}.

The missing data scenario is much less harmful to the model learned via the SOS-DVAE as compared to an SDVAE.
We show in Table~\ref{table:lfp-acc-auc-mc} that the SOS-DVAE outperforms the SDVAE at decoding behavioral context in all scenarios.
The single latent factor with the largest weight in the logistic regression classifier is shown in Fig. \ref{fig:electome_lfp_partial} next to the equivalent factor from a model with all channels present (Fig. \ref{fig:electome_lfp_full}).
We see that these factors share many features in common.

\begin{table}[h]\centering
    \caption{Prediction performance on LFP dataset with missing channels.\\}
    \resizebox{\columnwidth}{!}{
    
    \begin{tabular}{lccccc}
    
    \hline\hline
        &\multicolumn{2}{c}{MLP} & & \multicolumn{2}{c}{NMF} \\
        \cline{2-3} \cline{5-6}
        & ACC (\%) & AUC & & ACC (\%) & AUC\\
        
    \hline
        SDVAE& $75.12\pm.80$ & $.79\pm.017$ & & $70.95\pm1.39$ & $.74\pm.027$\\
        SOS-DVAE & $77.33\pm.46$ & $.81\pm.013$ & & $73.39\pm.02$ & $.78\pm.007$ \\
    \hline
    \end{tabular}
    }
    %}
     \label{table:lfp-acc-auc-mc}
\end{table}

\subsection{Additional results and comparison with other methods}
As can be seen from Algorithm \ref{alg:sosdvae}, $\mu$ weights the KL-divergence between the inference encoder $q_{\phi_1}(\cdot)$ and the classification $q_{\phi_2}(\cdot)$ encoder, with higher values for $\mu$ emphasizing that KL-divergence term over the classification loss. We tune $\mu$ to investigate the trade-off between prediction performance and the fidelity of generative model.  We report some of these results in the main paper, and included additional supplemental results here.

Figures \ref{fig:kl_curves}b, \ref{fig:kl_curves}d, and \ref{fig:kl_curves}f show the prediction performance against the KL-divergence on all three datasets using MLP decoders. We can see that the proposed SOS-DVAE obtains higher predictive performance (ACC or AUC) for the same level of KL-divergence.  As visualized in Figures \ref{fig:LFP_generative} for the LFP dataset and Figure \ref{fig:seed_generative} for the SEED dataset, a smaller KL-divergence (greater $\mu$) in SOD-DVAE and SDVAE means the $\phi_2$ encoders are more influenced by the generative model and thus are able to reconstruct the input sample much better. In both figures, the trained models with a Multilayer Perceptron (MLP) decoder are used to reconstruct an arbitrary input sample from the hold-out data. Note that the SVAE and SVAE-refit only has a single encoder thus they do not have a KL-divergence value.
The SVAE, SOS-DVAE, and SDVAE perform similarly in prediction. Similar to Figure \ref {fig:mnist_generative} for the MNIST dataset, the SVAE-refit presents a huge decrease in prediction. Despite predicting less well, the refitted model is able to generate the input features \textit{better} than that of SVAE.

As described in Section \ref{subsec:comparision_with_others_dgm} in the main manuscript, Figure 9 compares the generative ability to reconstruct an input image using VAE \citep{Kingma2013}, its semi-supervised extension \citep{kingma2014semi}, variational flows \citep{Rezende2015VariationalFlows}, auxiliary deep generative models \citep{maaloe2016auxiliary}, semi-supervised deep generative models with smooth-ELBO \citep{Feng_Kong_Chen_Zhang_Zhu_Chen_2021}, and the proposed SOS-VAE and SOS-DVAE. The methods in comparison are trained on the train set of MNIST using the default parameters in the publicly released code. The examples shown in Figure 9 are arbitrary unseen images from the test set of MNIST.

\begin{figure*}[h]
    \centering
    \includegraphics[width=0.97\textwidth]{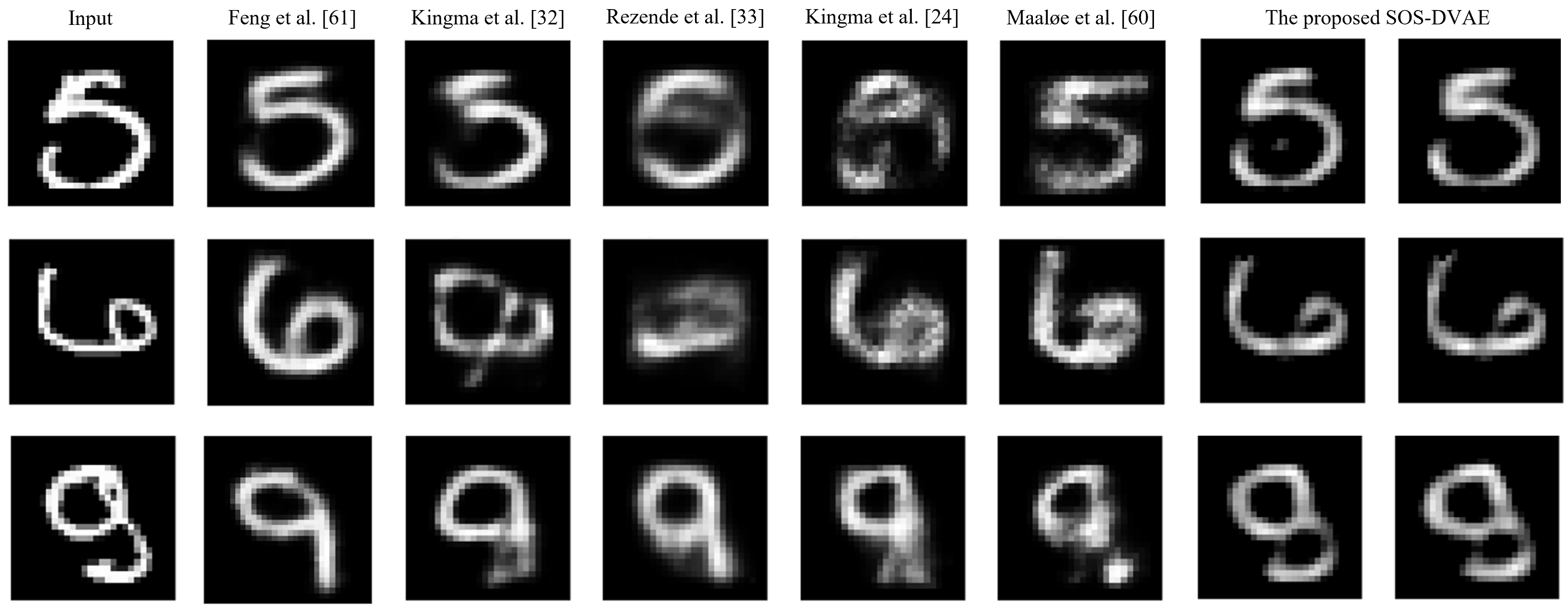}
    \caption{Demonstration of the promising generative ability of our models. The last 2 columns are from our proposed SOS-DVAE model, with the left column from the inference encoder, and the right column from the classification encoder. The KL-divergence between the 2 encoders is 0.5.}
    \label{fig:generative_compare}
\end{figure*}

\begin{figure*}[h]
\centering
\includegraphics[width=0.85\textwidth]{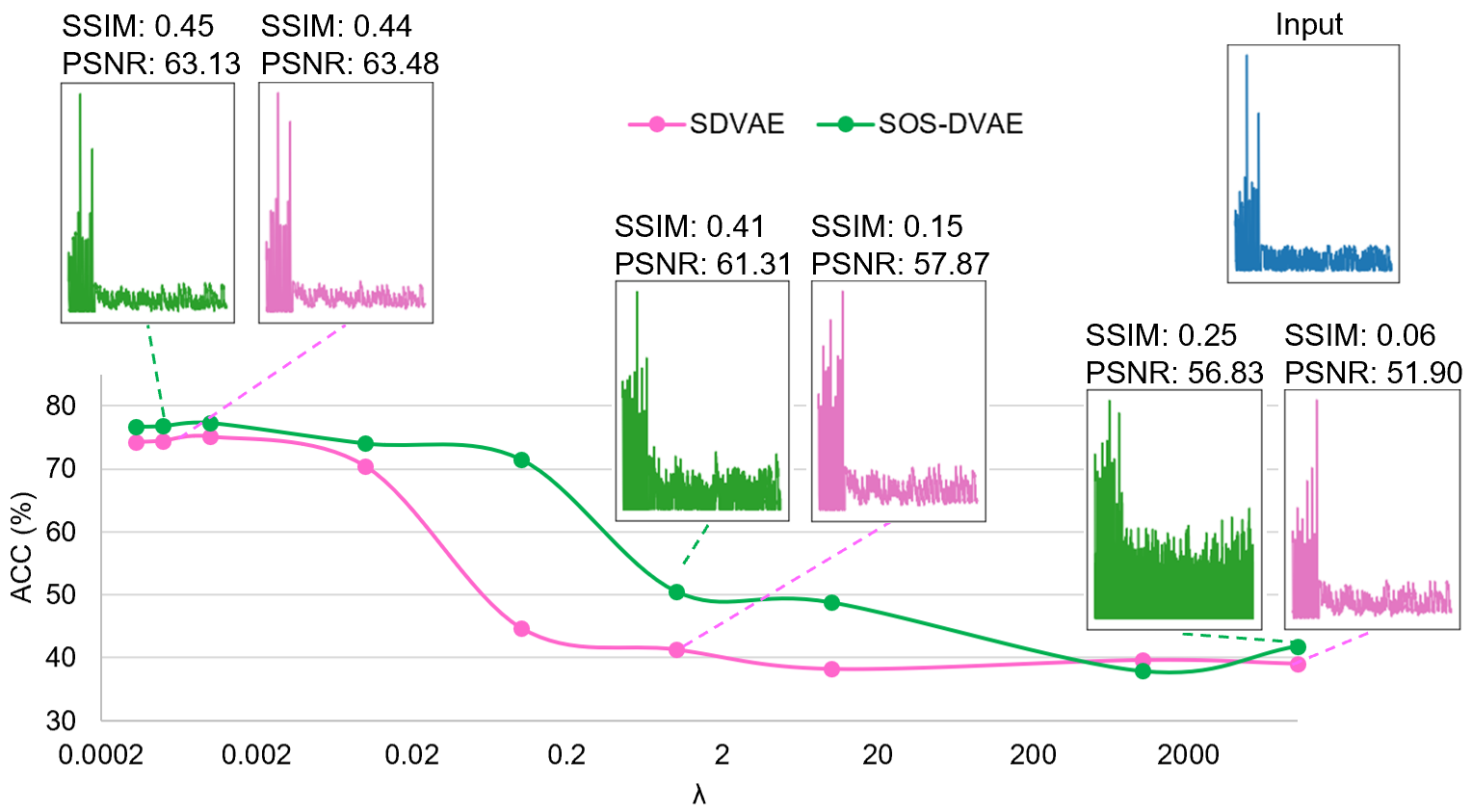}
\caption{Prediction and generation performance with the changing of $\lambda$ of LFP dataset with missing channel.}
\label{fig:tune_lambda_missing_lfp}
\end{figure*}

%% file: main.bbl
% Generated by IEEEtranN.bst, version: 1.14 (2015/08/26)
\begin{thebibliography}{75}
\providecommand{\natexlab}[1]{#1}
\providecommand{\url}[1]{#1}
\csname url@samestyle\endcsname
\providecommand{\newblock}{\relax}
\providecommand{\bibinfo}[2]{#2}
\providecommand{\BIBentrySTDinterwordspacing}{\spaceskip=0pt\relax}
\providecommand{\BIBentryALTinterwordstretchfactor}{4}
\providecommand{\BIBentryALTinterwordspacing}{\spaceskip=\fontdimen2\font plus
\BIBentryALTinterwordstretchfactor\fontdimen3\font minus
  \fontdimen4\font\relax}
\providecommand{\BIBforeignlanguage}[2]{{%
\expandafter\ifx\csname l@#1\endcsname\relax
\typeout{** WARNING: IEEEtranN.bst: No hyphenation pattern has been}%
\typeout{** loaded for the language `#1'. Using the pattern for}%
\typeout{** the default language instead.}%
\else
\language=\csname l@#1\endcsname
\fi
#2}}
\providecommand{\BIBdecl}{\relax}
\BIBdecl

\bibitem[Blei et~al.(2003)Blei, Ng, and Jordan]{Blei2003LatentAllocation}
D.~M. Blei, A.~Y. Ng, and M.~I. Jordan, ``Latent dirichlet allocation,''
  \emph{Journal of Machine Learning Research}, vol.~3, pp. 993--1022, 2003.

\bibitem[Perfors et~al.(2011)Perfors, Tenenbaum, Griffiths, and
  Xu]{Perfors2011ADevelopment}
A.~Perfors, J.~B. Tenenbaum, T.~L. Griffiths, and F.~Xu, ``{A tutorial
  introduction to Bayesian models of cognitive development},''
  \emph{Cognition}, vol. 120, no.~3, pp. 302--321, 2011.

\bibitem[Bonawitz and Griffiths(2010)]{Bonawitz2010DeconfoundingModels}
E.~Bonawitz and T.~L. Griffiths, ``{Deconfounding hypothesis generation and
  evaluation in Bayesian models},'' \emph{Proceedings of the Annual Meeting of
  the Cognitive Science Society}, vol.~32, no.~32, pp. 2260--2265, 2010.

\bibitem[Gallagher et~al.(2017)Gallagher, Ulrich, Talbot, Dzirasa, Carin, and
  Carlson]{Gallagher2017Cross-spectralAnalysis}
N.~Gallagher, K.~R. Ulrich, A.~Talbot, K.~Dzirasa, L.~Carin, and D.~E. Carlson,
  ``Cross-spectral factor analysis.'' in \emph{NeurIPS}, 2017, pp. 6842--6852.

\bibitem[Carlson et~al.(2017{\natexlab{a}})Carlson, David, Gallagher, Vu,
  Shirley, Hultman, Wang, Burrus, McClung, Kumar,
  et~al.]{Carlson2017Dynamically-timedPathway}
D.~Carlson, L.~K. David, N.~M. Gallagher, M.-A.~T. Vu, M.~Shirley, R.~Hultman,
  J.~Wang, C.~Burrus, C.~A. McClung, S.~Kumar \emph{et~al.}, ``Dynamically
  timed stimulation of corticolimbic circuitry activates a stress-compensatory
  pathway,'' \emph{Biological Psychiatry}, vol.~82, no.~12, pp. 904--913, 2017.

\bibitem[Hultman et~al.(2018)Hultman, Ulrich, Sachs, Blount, Carlson, Ndubuizu,
  Bagot, Parise, Vu, Gallagher, Wang, Silva, Deisseroth, Mague, Caron, Nestler,
  Carin, and Dzirasa]{Hultman2018Brain-wideVulnerability}
R.~Hultman, K.~Ulrich, B.~Sachs, C.~Blount, D.~Carlson, N.~Ndubuizu, R.~Bagot,
  E.~Parise, M.-A. Vu, N.~Gallagher, J.~Wang, A.~Silva, K.~Deisseroth,
  S.~Mague, M.~Caron, E.~Nestler, L.~Carin, and K.~Dzirasa, ``{Brain-wide
  electrical spatiotemporal dynamics encode depression vulnerability},''
  \emph{Cell}, vol. 173, no.~1, 2018.

\bibitem[Mague et~al.(2020{\natexlab{a}})Mague, Talbot, Blount, Duffney,
  Walder-Christensen, Adamson, Bey, Ndubuizu, Thomas, Hughes, Sinha, Fink,
  Gallagher, Fisher, Jiang, Carlson, and Dzirasa]{Mague2020Brain-wideState}
S.~D. Mague, A.~Talbot, C.~Blount, L.~J. Duffney, K.~K. Walder-Christensen,
  E.~Adamson, A.~L. Bey, N.~Ndubuizu, G.~Thomas, D.~N. Hughes, S.~Sinha, A.~M.
  Fink, N.~M. Gallagher, R.~L. Fisher, Y.-h. Jiang, D.~E. Carlson, and
  K.~Dzirasa, ``{Brain-wide electrical dynamics encode an appetitive
  socioemotional state},'' \emph{bioRxiv}, p. 2020.07.01.181347, 7 2020.

\bibitem[Block et~al.(2020)Block, Eroglu, Mague, Sriworarat, Blount, Malacon,
  Beben, Ndubuizu, Talbot, Gallagher, and {others}]{block2020prenatal}
C.~L. Block, O.~Eroglu, S.~D. Mague, C.~Sriworarat, C.~Blount, K.~E. Malacon,
  K.~A. Beben, N.~Ndubuizu, A.~Talbot, N.~Gallagher, and {others}, ``{Prenatal
  Environmental Stressors Impair Postnatal Microglia Function and Adult
  Behavior in Males},'' \emph{bioRxiv}, 2020.

\bibitem[Jolliffe(1982)]{Jolliffe1982}
I.~Jolliffe, ``{A note on the use of principal components in regression},''
  \emph{Journal of the Royal Statistical Society, Series C}, vol.~31, no.~3,
  pp. 300--303, 1982.

\bibitem[Mague et~al.(2020{\natexlab{b}})Mague, Talbot, Blount, Duffney,
  Walder-Christensen, Adamson, Bey, Ndubuizu, Thomas, Hughes, and
  {others}]{mague2020brain}
S.~D. Mague, A.~Talbot, C.~Blount, L.~J. Duffney, K.~K. Walder-Christensen,
  E.~Adamson, A.~L. Bey, N.~Ndubuizu, G.~Thomas, D.~N.~D. Hughes, and {others},
  ``{Brain-wide electrical dynamics encode an appetitive socioemotional
  state},'' \emph{bioRxiv}, 2020.

\bibitem[Jiang et~al.(2020)Jiang, Zhao, Qian, Song, and
  Lin]{jiang_generative_2020}
X.~Jiang, J.~Zhao, W.~Qian, W.~Song, and G.~N. Lin, ``A generative adversarial
  network model for disease gene prediction with rna-seq data,'' \emph{IEEE
  Access}, vol.~8, pp. 37\,352--37\,360, 2020.

\bibitem[Zhao et~al.(2017)Zhao, Dong, Chen, Iraji, Li, Makkie, Kou, and
  Liu]{zhao_constructing_2017}
Y.~Zhao, Q.~Dong, H.~Chen, A.~Iraji, Y.~Li, M.~Makkie, Z.~Kou, and T.~Liu,
  ``\BIBforeignlanguage{en}{Constructing fine-granularity functional brain
  network atlases via deep convolutional autoencoder},''
  \emph{\BIBforeignlanguage{en}{Medical Image Analysis}}, vol.~42, pp.
  200--211, 2017.

\bibitem[Vu et~al.(2018)Vu, Adalı, Ba, Buzs{\'{a}}ki, Carlson, Heller, Liston,
  Rudin, Sohal, Widge, Mayberg, Sapiro, and Dzirasa]{Vu2018ANeuroscience}
M.-A. Vu, T.~Adalı, D.~Ba, G.~Buzs{\'{a}}ki, D.~Carlson, K.~Heller, C.~Liston,
  C.~Rudin, V.~Sohal, A.~Widge, H.~Mayberg, G.~Sapiro, and K.~Dzirasa, ``{A
  shared vision for machine learning in neuroscience},'' \emph{Journal of
  Neuroscience}, vol.~38, no.~7, 2018.

\bibitem[Khorasani et~al.(2019)Khorasani, Shalchyan, and Daliri]{Khorasani2019}
A.~Khorasani, V.~Shalchyan, and M.~R. Daliri, ``{Adaptive artifact removal from
  intracortical channels for accurate decoding of a force signal in freely
  moving rats},'' \emph{Frontiers in Neuroscience}, vol.~13, pp. 1--12, 2019.

\bibitem[Joyce et~al.(2004)Joyce, Gorodnitsky, and Kutas]{Joyce2004}
C.~A. Joyce, I.~F. Gorodnitsky, and M.~Kutas, ``{Automatic removal of eye
  movement and blink artifactsfrom EEG data using blind component
  separation},'' \emph{Psychophysiology}, vol.~41, pp. 313--325, 2004.

\bibitem[Rasmus et~al.(2015)Rasmus, Valpola, Honkala, Berglund, and
  Raiko]{Rasmus2015Semi-SupervisedNetworks}
A.~Rasmus, H.~Valpola, M.~Honkala, M.~Berglund, and T.~Raiko, ``Semi-supervised
  learning with ladder networks,'' in \emph{NeurIPS}, 2015, pp. 3546--3554.

\bibitem[S{\o}nderby et~al.(2016)S{\o}nderby, Raiko, Maal{\o}e, S{\o}nderby,
  and Winther]{Raiko2018HowNetworks}
C.~K. S{\o}nderby, T.~Raiko, L.~Maal{\o}e, S.~K. S{\o}nderby, and O.~Winther,
  ``How to train deep variational autoencoders and probabilistic ladder
  networks,'' \emph{ICML}, vol.~48, 2016.

\bibitem[Pu et~al.(2016)Pu, Gan, Henao, Yuan, Li, Stevens, and Carin]{Pu2016}
Y.~Pu, Z.~Gan, R.~Henao, X.~Yuan, C.~Li, A.~Stevens, and L.~Carin,
  ``{Variational autoencoder for deep learning of images, labels and
  captions},'' in \emph{NeurIPS}, 2016, pp. 2360--2368.

\bibitem[Le et~al.(2018)Le, Patterson, and White]{Le2018}
L.~Le, A.~Patterson, and M.~White, ``{Supervised autoencoders: improving
  generalization performance with unsupervised regularizers},'' in
  \emph{NeurIPS}, 2018, pp. 107--117.

\bibitem[Finn et~al.(2017)Finn, Abbeel, and
  Levine]{Finn2017Model-agnosticNetworks}
C.~Finn, P.~Abbeel, and S.~Levine, ``{Model-agnostic meta-learning for fast
  adaptation of deep networks},'' in \emph{ICML}, 2017, pp. 1856--1868.

\bibitem[Liu et~al.(2019)Liu, Davison, and
  Johns]{Liu2019Self-SupervisedLearning}
S.~Liu, A.~J. Davison, and E.~Johns, ``Self-supervised generalisation with meta
  auxiliary learning,'' \emph{NeurIPS}, 2019.

\bibitem[Plummer(2014)]{Plummer2014}
M.~Plummer, ``{Cuts in Bayesian graphical models},'' \emph{Statistics and
  Computing}, vol.~25, no.~1, pp. 37--43, 2014.

\bibitem[Hahn et~al.(2013)Hahn, Carvalho, and Mukherjee]{Hahn2013}
P.~R. Hahn, C.~M. Carvalho, and S.~Mukherjee, ``{Partial factor modeling:
  predictor-dependent shrinkage for linear regression},'' \emph{Journal of the
  American Statistical Association}, vol. 108, no. 503, pp. 999--1008, 2013.

\bibitem[Kingma and Welling(2013)]{Kingma2013}
D.~P. Kingma and M.~Welling, ``{Auto-encoding variational Bayes},'' in
  \emph{ICLR}, 2013.

\bibitem[Yu et~al.(2006)Yu, Yu, Tresp, Kriegel, and Wu]{Yu2006}
S.~Yu, K.~Yu, V.~Tresp, H.~P. Kriegel, and M.~Wu, ``{Supervised probabilistic
  principal component analysis},'' in \emph{ACM SIGKDD}, vol. 2006, 2006, pp.
  464--473.

\bibitem[Gao et~al.(2011)Gao, Wang, Tao, and Li]{Gao2011SupervisedReduction}
X.~Gao, X.~Wang, D.~Tao, and X.~Li, ``{Supervised Gaussian process latent
  variable model for dimensionality reduction},'' \emph{IEEE T. Systems, Man,
  and Cybernetics}, vol.~41, no.~2, pp. 425--434, 2011.

\bibitem[Bhattacharya and Dunson(2011)]{Bhattacharya2011}
A.~Bhattacharya and D.~B. Dunson, ``{Sparse Bayesian infinite factor models},''
  \emph{Biometrika}, vol.~98, no.~2, pp. 291--306, 2011.

\bibitem[Talbot et~al.(2020)Talbot, Dunson, Dzirasa, and
  Carlson]{Talbot2020SupervisedActivity}
A.~Talbot, D.~Dunson, K.~Dzirasa, and D.~Carlson, ``Supervised autoencoders
  learn robust joint factor models of neural activity,'' \emph{arXiv preprint
  arXiv:2004.05209}, 2020.

\bibitem[Parthasarathy and Busso(2018)]{Parthasarathy2018LadderAttributes}
S.~Parthasarathy and C.~Busso, ``Ladder networks for emotion recognition: Using
  unsupervised auxiliary tasks to improve predictions of emotional
  attributes,'' \emph{Conference of the International Speech Communication
  Association, INTERSPEECH}, pp. 3698--3702, 2018.

\bibitem[Liebel and K{\"o}rner(2018)]{Liebel2018AuxiliaryLearning}
L.~Liebel and M.~K{\"o}rner, ``Auxiliary tasks in multi-task learning,''
  \emph{arXiv preprint arXiv:1805.06334}, 2018.

\bibitem[Zhao et~al.(2019)Zhao, Zhang, Zhang, and
  Zhang]{Zhao2019LeveragingCounting}
M.~Zhao, J.~Zhang, C.~Zhang, and W.~Zhang, ``Leveraging heterogeneous auxiliary
  tasks to assist crowd counting,'' in \emph{IEEE/CVF CVPR}, 2019, pp.
  12\,736--12\,745.

\bibitem[Kingma et~al.(2014)Kingma, Rezende, Mohamed, and
  Welling]{kingma2014semi}
D.~P. Kingma, D.~J. Rezende, S.~Mohamed, and M.~Welling, ``Semi-supervised
  learning with deep generative models,'' \emph{NeurIPS}, 2014.

\bibitem[Rezende and Mohamed(2015)]{Rezende2015VariationalFlows}
D.~J. Rezende and S.~Mohamed, ``{Variational inference with normalizing
  flows},'' vol.~37, 2015.

\bibitem[Joy et~al.(2021)Joy, Schmon, Torr, Siddharth, and
  Rainforth]{joy2021capturing}
T.~Joy, S.~M. Schmon, P.~H.~S. Torr, N.~Siddharth, and T.~Rainforth,
  ``Capturing label characteristics in vaes,'' 2021.

\bibitem[Mueller et~al.(2017)Mueller, Gifford, and
  Jaakkola]{pmlr-v70-mueller17a}
J.~Mueller, D.~Gifford, and T.~Jaakkola, ``Sequence to better sequence:
  continuous revision of combinatorial structures,'' \emph{ICML}, vol.~70, pp.
  2536--2544, 2017.

\bibitem[Andrychowicz et~al.(2016)Andrychowicz, Denil, Colmenarejo, Hoffman,
  Pfau, Schaul, Shillingford, and de~Freitas]{Andrychowicz2016LearningDescent}
M.~Andrychowicz, M.~Denil, S.~G. Colmenarejo, M.~W. Hoffman, D.~Pfau,
  T.~Schaul, B.~Shillingford, and N.~de~Freitas, ``Learning to learn by
  gradient descent by gradient descent,'' in \emph{NeurIPS}, 2016, pp.
  3988--3996.

\bibitem[Carlson et~al.(2017{\natexlab{b}})Carlson, David, Gallagher, Vu,
  Shirley, Hultman, Wang, Burrus, McClung, Kumar, Carin, Mague, and
  Dzirasa]{Carlson2017}
D.~Carlson, L.~K. David, N.~M. Gallagher, M.~A.~T. Vu, M.~Shirley, R.~Hultman,
  J.~Wang, C.~Burrus, C.~A. McClung, S.~Kumar, L.~Carin, S.~D. Mague, and
  K.~Dzirasa, ``{Dynamically Timed Stimulation of Corticolimbic Circuitry
  Activates a Stress-Compensatory Pathway},'' pp. 904--913, 2017.

\bibitem[Dempe et~al.(2015)Dempe, Kalashnikov, P{\'e}rez-Vald{\'e}s, and
  Kalashnykova]{dempe2015bilevel}
S.~Dempe, V.~Kalashnikov, G.~A. P{\'e}rez-Vald{\'e}s, and N.~Kalashnykova,
  ``Bilevel programming problems,'' \emph{Energy Systems. Springer, Berlin},
  vol.~10, pp. 978--3, 2015.

\bibitem[Dempe et~al.(2007)Dempe, Dutta, and Mordukhovich]{dempe2007new}
S.~Dempe, J.~Dutta, and B.~Mordukhovich, ``New necessary optimality conditions
  in optimistic bilevel programming,'' \emph{Optimization}, vol.~56, no. 5-6,
  pp. 577--604, 2007.

\bibitem[Humpherys and Jarvis(2021)]{humpherys2021foundations}
J.~Humpherys and T.~J. Jarvis, \emph{Foundations of Applied Mathematics, Volume
  II: Algorithms,Approximation, Optimization}.\hskip 1em plus 0.5em minus
  0.4em\relax SIAM, 2021, vol. 152.

\bibitem[Dempe and Zemkoho(2012)]{dempe2012karush}
S.~Dempe and A.~B. Zemkoho, ``On the karush--kuhn--tucker reformulation of the
  bilevel optimization problem,'' \emph{Nonlinear Analysis: Theory, Methods \&
  Applications}, vol.~75, no.~3, pp. 1202--1218, 2012.

\bibitem[Mague et~al.(2022)Mague, Talbot, Blount, Walder-Christensen, Duffney,
  Adamson, Bey, Ndubuizu, Thomas, Hughes, et~al.]{mague2022brain}
S.~D. Mague, A.~Talbot, C.~Blount, K.~K. Walder-Christensen, L.~J. Duffney,
  E.~Adamson, A.~L. Bey, N.~Ndubuizu, G.~E. Thomas, D.~N. Hughes \emph{et~al.},
  ``Brain-wide electrical dynamics encode individual appetitive social
  behavior,'' \emph{Neuron}, 2022.

\bibitem[Mirrlees(1999)]{mirrlees1999theory}
J.~A. Mirrlees, ``The theory of moral hazard and unobservable behaviour: Part
  i,'' \emph{The Review of Economic Studies}, vol.~66, no.~1, pp. 3--21, 1999.

\bibitem[Jacot et~al.(2018)Jacot, Gabriel, and Hongler]{jacot2018neural}
A.~Jacot, F.~Gabriel, and C.~Hongler, ``Neural tangent kernel: Convergence and
  generalization in neural networks,'' \emph{Advances in neural information
  processing systems}, vol.~31, 2018.

\bibitem[Allen-Zhu et~al.(2019)Allen-Zhu, Li, and Song]{allen2019convergence}
Z.~Allen-Zhu, Y.~Li, and Z.~Song, ``A convergence theory for deep learning via
  over-parameterization,'' in \emph{International Conference on Machine
  Learning}.\hskip 1em plus 0.5em minus 0.4em\relax PMLR, 2019, pp. 242--252.

\bibitem[Ergen and Pilanci(2021)]{ergen2021convex}
T.~Ergen and M.~Pilanci, ``Convex geometry and duality of over-parameterized
  neural networks,'' \emph{Journal of Machine Learning Research}, vol.~22, pp.
  1--63, 2021.

\bibitem[Draxler et~al.(2018)Draxler, Veschgini, Salmhofer, and
  Hamprecht]{draxler2018essentially}
F.~Draxler, K.~Veschgini, M.~Salmhofer, and F.~Hamprecht, ``Essentially no
  barriers in neural network energy landscape,'' in \emph{International
  conference on machine learning}.\hskip 1em plus 0.5em minus 0.4em\relax PMLR,
  2018, pp. 1309--1318.

\bibitem[Jin et~al.(2017)Jin, Ge, Netrapalli, Kakade, and
  Jordan]{jin2017escape}
C.~Jin, R.~Ge, P.~Netrapalli, S.~M. Kakade, and M.~I. Jordan, ``How to escape
  saddle points efficiently,'' in \emph{International Conference on Machine
  Learning}.\hskip 1em plus 0.5em minus 0.4em\relax PMLR, 2017, pp. 1724--1732.

\bibitem[Humpherys et~al.(2017)Humpherys, Jarvis, and
  Evans]{humpherys2017foundations}
J.~Humpherys, T.~J. Jarvis, and E.~J. Evans, \emph{Foundations of Applied
  Mathematics, Volume I: Mathematical Analysis}.\hskip 1em plus 0.5em minus
  0.4em\relax SIAM, 2017, vol. 152.

\bibitem[Fallah et~al.(2020)Fallah, Mokhtari, and
  Ozdaglar]{fallah2020convergence}
A.~Fallah, A.~Mokhtari, and A.~Ozdaglar, ``On the convergence theory of
  gradient-based model-agnostic meta-learning algorithms,'' in
  \emph{International Conference on Artificial Intelligence and
  Statistics}.\hskip 1em plus 0.5em minus 0.4em\relax PMLR, 2020, pp.
  1082--1092.

\bibitem[Armstrong et~al.(2013)Armstrong, Krook-Magnuson, Oijala, and
  Soltesz]{armstrong2013closed}
C.~Armstrong, E.~Krook-Magnuson, M.~Oijala, and I.~Soltesz, ``Closed-loop
  optogenetic intervention in mice,'' \emph{Nature protocols}, vol.~8, no.~8,
  pp. 1475--1493, 2013.

\bibitem[Basu et~al.(1998)Basu, Harris, Hjort, and Jones]{basu1998robust}
A.~Basu, I.~R. Harris, N.~L. Hjort, and M.~Jones, ``Robust and efficient
  estimation by minimising a density power divergence,'' \emph{Biometrika},
  vol.~85, no.~3, pp. 549--559, 1998.

\bibitem[Van~Erven and Harremos(2014)]{van2014renyi}
T.~Van~Erven and P.~Harremos, ``R{\'e}nyi divergence and kullback-leibler
  divergence,'' \emph{IEEE Transactions on Information Theory}, vol.~60, no.~7,
  pp. 3797--3820, 2014.

\bibitem[M{\"u}ller(1997)]{muller1997integral}
A.~M{\"u}ller, ``Integral probability metrics and their generating classes of
  functions,'' \emph{Advances in Applied Probability}, vol.~29, no.~2, pp.
  429--443, 1997.

\bibitem[Khambhati et~al.(2018)Khambhati, Sizemore, Betzel, and
  Bassett]{khambhati2018modeling}
A.~N. Khambhati, A.~E. Sizemore, R.~F. Betzel, and D.~S. Bassett, ``Modeling
  and interpreting mesoscale network dynamics,'' \emph{Neuroimage}, vol. 180,
  pp. 337--349, 2018.

\bibitem[Medaglia et~al.(2015)Medaglia, Lynall, and Bassett]{Medaglia2015}
J.~D. Medaglia, M.~E. Lynall, and D.~S. Bassett, ``{Cognitive network
  neuroscience},'' \emph{Journal of Cognitive Neuroscience}, vol.~27, no.~8,
  pp. 1471--1491, 2015.

\bibitem[Deng(2012)]{deng2012mnist}
L.~Deng, ``The mnist database of handwritten digit images for machine learning
  research [best of the web],'' \emph{IEEE Signal Processing Magazine},
  vol.~29, no.~6, pp. 141--142, 2012.

\bibitem[Heusel et~al.(2017)Heusel, Ramsauer, Unterthiner, Nessler, and
  Hochreiter]{heusel2017gans}
M.~Heusel, H.~Ramsauer, T.~Unterthiner, B.~Nessler, and S.~Hochreiter, ``Gans
  trained by a two time-scale update rule converge to a local nash
  equilibrium,'' in \emph{NeurIPS}, 2017, pp. 6629--6640.

\bibitem[Goodfellow et~al.(2014)Goodfellow, Pouget-Abadie, Mirza, Xu,
  Warde-Farley, Ozair, Courville, and Bengio]{Goodfellow2014GenerativeNets}
I.~Goodfellow, J.~Pouget-Abadie, M.~Mirza, B.~Xu, D.~Warde-Farley, S.~Ozair,
  A.~Courville, and Y.~Bengio, ``{Generative adversarial nets},'' \emph{NeurIPS
  27}, pp. 2672--2680, 2014.

\bibitem[Maal{\o}e et~al.(2016)Maal{\o}e, S{\o}nderby, S{\o}nderby, and
  Winther]{maaloe2016auxiliary}
L.~Maal{\o}e, C.~K. S{\o}nderby, S.~K. S{\o}nderby, and O.~Winther, ``Auxiliary
  deep generative models,'' in \emph{ICML}, 2016, pp. 1445--1453.

\bibitem[Feng et~al.(2021)Feng, Kong, Chen, Zhang, Zhu, and
  Chen]{Feng_Kong_Chen_Zhang_Zhu_Chen_2021}
H.-Z. Feng, K.~Kong, M.~Chen, T.~Zhang, M.~Zhu, and W.~Chen, ``Shot-vae:
  semi-supervised deep generative models with label-aware elbo
  approximations,'' \emph{AAAI}, vol.~35, no.~8, pp. 7413--7421, May 2021.

\bibitem[Caruana(1997)]{caruana1997multitask}
R.~Caruana, ``Multitask learning,'' \emph{Machine learning}, vol.~28, no.~1,
  pp. 41--75, 1997.

\bibitem[Duan et~al.(2013)Duan, Zhu, and Lu]{duan2013differential}
R.-N. Duan, J.-Y. Zhu, and B.-L. Lu, ``Differential entropy feature for
  {EEG}-based emotion classification,'' in \emph{IEEE/EMBS NER}.\hskip 1em plus
  0.5em minus 0.4em\relax IEEE, 2013, pp. 81--84.

\bibitem[Zheng and Lu(2015)]{zheng2015investigating}
W.-L. Zheng and B.-L. Lu, ``Investigating critical frequency bands and channels
  for {EEG}-based emotion recognition with deep neural networks,'' \emph{IEEE
  Transactions on Autonomous Mental Development}, vol.~7, no.~3, pp. 162--175,
  2015.

\bibitem[Wong(2015)]{wong_performance_2015}
T.-T. Wong, ``\BIBforeignlanguage{en}{Performance evaluation of classification
  algorithms by k-fold and leave-one-out cross validation},''
  \emph{\BIBforeignlanguage{en}{Pattern Recognition}}, vol.~48, no.~9, pp.
  2839--2846, 2015.

\bibitem[Rudin(2019)]{Rudin2019StopInstead}
C.~Rudin, ``{Stop explaining black box machine learning models for high stakes
  decisions and use interpretable models instead},'' \emph{Nature Machine
  Intelligence}, vol.~1, no.~5, pp. 206--215, 5 2019.

\bibitem[Rudin and Carlson(2019)]{Rudin2019TheAnalysisc}
C.~Rudin and D.~Carlson, ``{The secrets of machine learning: ten things you
  wish you had known earlier to be more effective at data analysis},'' in
  \emph{Operations Research {\&} Management Science in the Age of
  Analytics}.\hskip 1em plus 0.5em minus 0.4em\relax INFORMS, 10 2019, pp.
  44--72.

\bibitem[Li et~al.(2018)Li, Zhu, and Zhang]{li2017max}
C.~Li, J.~Zhu, and B.~Zhang, ``Max-margin deep generative models for (semi-)
  supervised learning,'' \emph{IEEE TPAMI}, vol.~40, no.~11, pp. 2762--2775,
  2018.

\bibitem[Ranganath et~al.(2014)Ranganath, Gerrish, and
  Blei]{Ranganath2014BlackInference}
R.~Ranganath, S.~Gerrish, and D.~M. Blei, ``{Black box variational
  inference},'' in \emph{Journal of Machine Learning Research}, vol.~33, 2014,
  pp. 814--822.

\bibitem[Tran et~al.(2016)Tran, Kucukelbir, Dieng, Rudolph, Liang, and
  Blei]{Tran2017Edward:Criticism}
D.~Tran, A.~Kucukelbir, A.~B. Dieng, M.~Rudolph, D.~Liang, and D.~M. Blei,
  ``Edward: A library for probabilistic modeling, inference, and criticism,''
  \emph{arXiv preprint arXiv:1610.09787}, 2016.

\bibitem[Dillon et~al.(2017)Dillon, Langmore, Tran, Brevdo, Vasudevan, Moore,
  Patton, Alemi, Hoffman, and Saurous]{Dillon2017TensorFlowDistributions}
J.~V. Dillon, I.~Langmore, D.~Tran, E.~Brevdo, S.~Vasudevan, D.~Moore,
  B.~Patton, A.~Alemi, M.~Hoffman, and R.~A. Saurous, ``{TensorFlow
  distributions},'' \emph{arXiv preprint arXiv:1711.10604}, 2017.

\bibitem[Soudry et~al.(2015)Soudry, Keshri, Stinson, Oh, Iyengar, and
  Paninski]{soudry2015efficient}
D.~Soudry, S.~Keshri, P.~Stinson, M.-h. Oh, G.~Iyengar, and L.~Paninski,
  ``Efficient "shotgun" inference of neural connectivity from highly
  sub-sampled activity data,'' \emph{PLoS Comput Biol}, vol.~11, no.~10, p.
  e1004464, 2015.

\bibitem[Gelman et~al.(1995)Gelman, Carlin, Stern, and
  Rubin]{gelman1995bayesian}
A.~Gelman, J.~B. Carlin, H.~S. Stern, and D.~B. Rubin, \emph{Bayesian data
  analysis}.\hskip 1em plus 0.5em minus 0.4em\relax Chapman and Hall/CRC, 1995.

\bibitem[Welch(1967)]{Welch1967ThePeriodograms}
P.~Welch, ``{The use of fast Fourier transform for the estimation of power
  spectra: A method based on time averaging over short, modified
  periodograms},'' \emph{IEEE T. Audio Electro.}, vol.~15, no.~2, pp. 70--73, 6
  1967.

\bibitem[Rabiner and Gold(1975)]{Rabiner1975}
L.~Rabiner and B.~Gold, \emph{{Theory and Application of Digital Signal
  Processing}}.\hskip 1em plus 0.5em minus 0.4em\relax Englewood Cliffs, NJ:
  Prentice-Hall, 1975.

\end{thebibliography}
